\documentclass[10pt,journal,compsoc]{IEEEtran}
\ifCLASSOPTIONcompsoc
  \usepackage[nocompress]{cite}
\else
  \usepackage{cite}
\fi
\ifCLASSINFOpdf
  \usepackage[pdftex]{graphicx}
\else
\fi
\usepackage{amsfonts}
\usepackage{amsmath}

\usepackage{url}

\usepackage{chngcntr}
\usepackage[resetlabels]{multibib}
\newcites{append}{References}

\newtheorem{problem}{Problem}

\begin{document}
%
\title{Continuously Controllable Facial Expression Editing in Talking Face Videos}
%
%
%
%

\author{Zhiyao Sun,
        Yu-Hui Wen,
        Tian Lv,
        Yanan Sun,
        Ziyang Zhang,
        Yaoyuan Wang,
        and Yong-Jin Liu%
\IEEEcompsocitemizethanks{\IEEEcompsocthanksitem Z. Sun, T. Lv, Y. Sun and Y. Liu are
with BNRist, the Department of Computer Science and Technology, MOE-Key Laboratory of Pervasive Computing, Tsinghua University, Beijing 100084, China.\protect\\
E-mail: \{sunzy21@mails., lt22@mails., sunyn20@mails., liuyongjin@\}tsinghua.edu.cn
\IEEEcompsocthanksitem Y. Wen is with the Beijing Key Laboratory of Traffic Data Analysis and Mining, School of Computer and Information Technology, Beijing Jiaotong University, Beijing 100044, China.\protect\\
E-mail: yhwen1@bjtu.edu.cn
\IEEEcompsocthanksitem Z. Zhang and Y. Wang are with the Advanced Computing and Storage Lab, Huawei Technologies Company Ltd., Shenzhen 518129, China.\protect\\
E-mail: \{zhangziyang11, wangyaoyuan1\}@huawei.com}
\thanks{Corresponding authors: Yong-Jin Liu and Yu-Hui Wen.}
\thanks{This work was partially supported by National Key R\&D Program of China (2022ZD0117900), the Natural Science Foundation of China (62332019, 62202257, U2336214) and Beijing Jiaotong University Youth Elite Project (2023XKRC045).}
}

%
%

\markboth{To appear in IEEE Transactions on Affective Computing. DOI: 10.1109/TAFFC.2023.3334511.}%
{Sun \MakeLowercase{\textit{et al.}}: Continuously Controllable Facial Expression Editing in Talking Face Videos}
%



\IEEEtitleabstractindextext{%
\begin{abstract}
Recently audio-driven talking face video generation has attracted considerable attention. However, very few researches address the issue of emotional editing of these talking face videos with continuously controllable expressions, which is a strong demand in the industry. The challenge is that speech-related expressions and emotion-related expressions are often highly coupled. Meanwhile, traditional image-to-image translation methods cannot work well in our application due to the coupling of expressions with other attributes such as poses, i.e., translating the expression of the character in each frame may simultaneously change the head pose due to the bias of the training data distribution. In this paper, we propose a high-quality facial expression editing method for talking face videos, allowing the user to control the target emotion in the edited video continuously.
We present a new perspective for this task as a special case of motion information editing, where we use a 3DMM to capture major facial movements and an associated texture map modeled by a StyleGAN to capture appearance details. Both representations (3DMM and texture map) contain emotional information and can be continuously modified by neural networks and easily smoothed by averaging in coefficient/latent spaces, making our method simple yet effective. We also introduce a mouth shape preservation loss to control the trade-off between lip synchronization and the degree of exaggeration of the edited expression. Extensive experiments and a user study show that our method achieves state-of-the-art performance across various evaluation criteria.

\end{abstract}

\begin{IEEEkeywords}
Facial expression editing, continuously controllable expressions, talking face video generation.
\end{IEEEkeywords}}

\maketitle

\IEEEdisplaynontitleabstractindextext

%
\IEEEpeerreviewmaketitle

\IEEEraisesectionheading{\section{Introduction}\label{sec:introduction}}

%
%
%
%

\begin{figure*}[!ht]
    \centering
    \includegraphics[width=\textwidth]{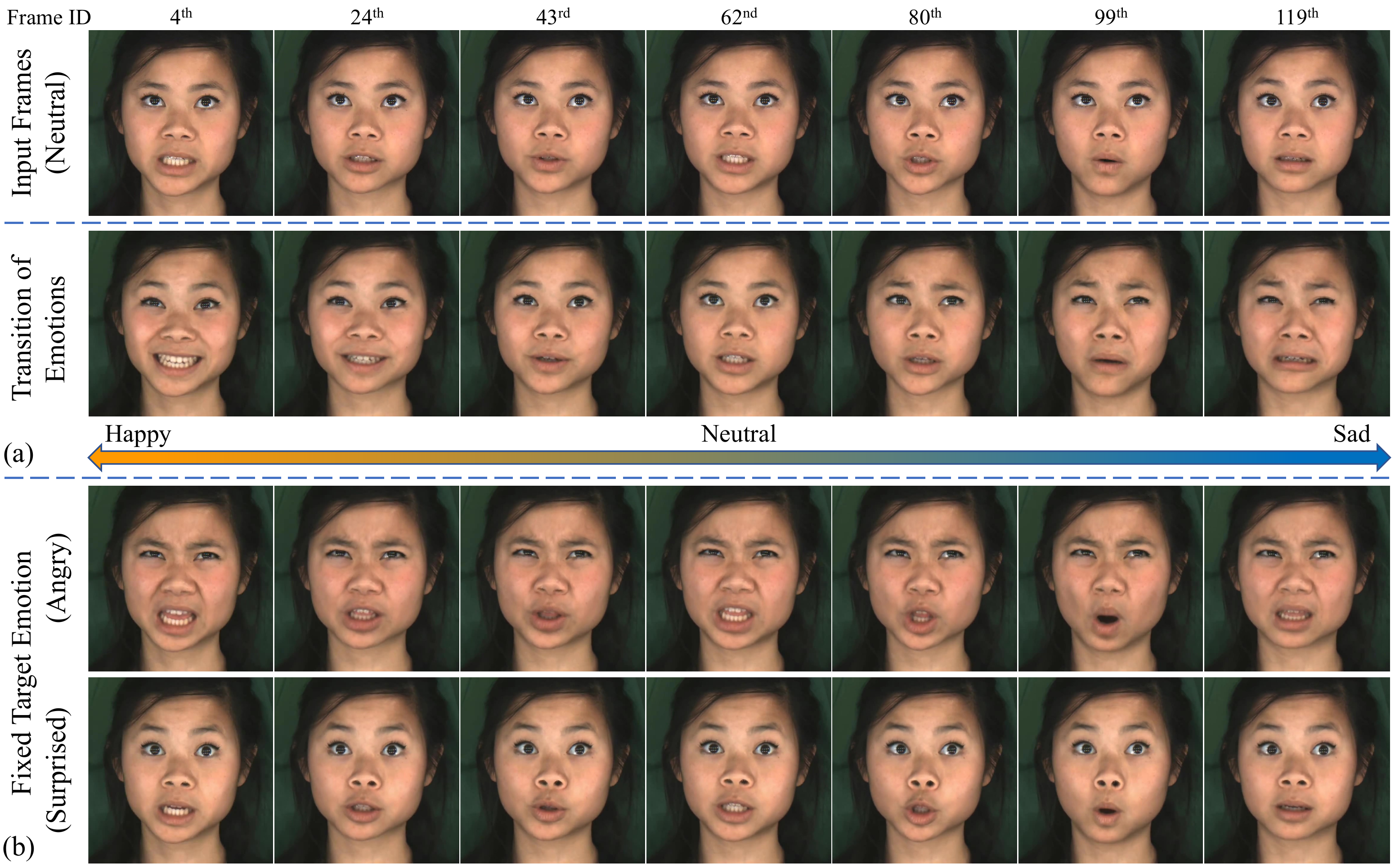}
    \caption{
        Our method takes a video with neutral facial expression (top row) and target emotion labels (including emotion type and intensity) of each frame as input and outputs the corresponding emotional video. (a) Our method can input continuously changing intensity values and generate smooth transitions between different emotions. (b) Our method can also input fixed expression type and intensity value and generate the corresponding emotional video. Animation effects can be found in the supplementary demo video.
    }
    \label{fig:overview}
\end{figure*}

\IEEEPARstart{I}{n} recent years, numerous methods have been proposed to generate realistic-looking talking videos driven by audio input, e.g., \cite{chen2019hierarchical, thies2020neural, vougioukas2020realistic, yi2020audio, prajwal2020lip, lahiri2021lipsync3d}. Much effort has been made to improve the quality of the generated result, mainly on identity preservation and lip synchronization. However, most existing studies have not explicitly taken emotions into consideration and cannot explicitly control the generated expression. A few recent works \cite{ji2021audio, ye2022dynamic,papantoniou22ned} have attempted to generate talking videos with pre-specified emotions; however, the accuracy and naturalness of the generated expression and the lip synchronization are not satisfactory due to the coupled nature of the speech-related facial expression and the emotion-related facial expression, which makes the continuously controllable facial expression generation problem highly challenging.

Our study is based on the following key observation: Since directly generating talking videos with realistic and controllable expressions is very challenging, we decompose this task into two sub-tasks, each of which can find a reasonable solution: (1) generating an emotionless talking face video and (2) editing the expression in each frame using pre-specified emotion labels. Many existing works can already do the first sub-task well. In this paper, we propose a simple yet effective method to accomplish the second sub-task (Figure \ref{fig:overview}). 

Intuitively, existing image-to-image translation methods (e.g., \cite{isola2017image, zhu2017unpaired, choi2018stargan}) or facial attribute editing methods (e.g., \cite{shen2020interpreting,tripathy20icface}) could potentially be used for the second sub-task. However, our preliminary study shows that these methods do not work well for three reasons. First, these methods may change the shape of the mouth, making the edited video out of sync with the audio. Second, image-to-image translation methods may simultaneously change the pose, lighting, or other attributes if they are entangled with the facial expressions in the training data, leading to problems such as misalignment. We also pay attention to the StyleGAN model \cite{karras2019style, karras2020analyzing}, which can generate high-quality images and learn semantically disentangled latent space. Many state-of-the-art studies \cite{shen2020interpreting, richardson2021encoding, tov2021designing, alaluf2021restyle} have explored its powerful editing capabilities. However, finding distinguishable and decoupled editing directions for different expressions is still challenging. Third, applying these image translation/editing methods to a video frame-by-frame may cause a serious inter-frame discontinuity problem.

In this paper, we tackle the second sub-task from a new perspective, that is, we regard expression editing in a talking video as a specific type of motion information editing. This perspective is based on an observation that when a person speaks, his/her facial muscles control both expressions and mouth movements. Accordingly, our method uses a two-level motion information approach for expression editing: (1) a 3DMM \cite{blanz1999morphable} to model the primary facial movements and (2) a deliberate texture map generated by StyleGAN \cite{karras2020analyzing} to capture high-quality appearance details and their subtle variations. Our experiments find that the 3DMM is able to capture large-scale deformations such as opening the mouth wide in anger or raising eyebrows in joy, which can impact people's perception of whether an expression is positive or negative. However, we observe that due to limited expressiveness, 3DMM cannot capture color changes and some fine facial details such as wrinkles. These details affect people's perception of subtle facial expressions such as sadness, anger, fear, disgust, and contempt in negative emotions. Therefore, we incorporate a high-quality texture map to capture these details. Given this two-level motion information representation, our expression editing method first reconstructs the 3D face and extracts the texture from the input frame. Next, our method transforms the 3DMM shape coefficient and the texture of the 3D face to the target emotion, respectively. Finally, our method renders the face image using the reconstructed pose, the generated 3D face shape and texture.

Our two-level motion information editing strategy is effective and simple to use and has the following advantages. First, since we decompose the facial expression into a 3DMM shape and an associate texture map, we can focus on the pose-independent shape and texture respectively during expression editing. This enables our method to be unaffected by the coupling of emotions and poses from the bias of the training data distribution. Second, with the aid of the key points in the 3DMM, we can use a simple loss function to impose an explicit constraint on the mouth shape, giving us control over the tradeoff between mouth shape preservation and the degree of exaggeration in the edited expression. Third, our method does not require complex network structures or loss functions to ensure inter-frame continuity. By decoupling the motion information into major facial movement (via 3DMM) and subtle changes in fine appearance details (via texture map), we can smooth major shape information and fine appearance details separately in adjacent frames and then combine them to achieve inter-frame continuity. As the 3DMM we used is a linear model, we can smooth the face shape by simply taking the average of 3DMM coefficients inferred from adjacent frames. Appearance details can also be smoothed easily by averaging the codes in the latent space of textures modeled by StyleGAN. Fourth, our use of textures generated by StyleGAN enables high-quality editing at resolutions up to $1024\times1024$.

We evaluate our proposed method comprehensively and compare it with some existing baseline methods. Some commonly used metrics for lip synchronization, identity preservation and image quality of the generated results are adopted in our evaluation, including LSE-D and LSE-C, the average ArcFace feature, FID and CPBD, etc.; see Section \ref{subsec:metrics} for details. In addition, we adopt the FED metric to assess the reality degree of the generated expressions and propose a novel metric called the LIE metric. This metric is specifically designed to evaluate the change rate in the intensity of facial expression editing, which can indicate the smoothness of the method's control over facial expressions.

In summary, we make three contributions in this paper:

\begin{itemize}
  \item We characterize the facial expression editing problem as a special motion information editing problem and propose a two-level expression representation to decompose the motion information into major facial movement (by 3DMM) and subtle changes in fine appearance details (by texture map). 
  
  \item Based on the two-level expression representation, we propose a simple yet effective method for continuously controllable facial expression editing in talking face videos, with delicate choices of architectures and losses to enable accurate and high-resolutional expression editing while preserving identity, lip synchronization and temporal consistency.
  
  \item We propose a new metric --- LIE, to evaluate the linearity of intensity editing in the scenario of expression editing. 
  
\end{itemize}

\section{Related Work}

\subsection{3D Face Model and Reconstruction}

Parametric models are popular for 3D face representation. The 3D Morphable Face Model (3DMM) \cite{blanz1999morphable} is probably the most popular 3D parametric face model, which uses shape and texture vectors to characterize the statistical distribution of face shape and textures. Many variants of 3DMMs have been proposed (e.g., \cite{paysan20093d, cao2013facewarehouse, li2017learning}), most of which are linear models built by performing principal component analysis on a set of scanned 3D face data. The Basel Face Model (BFM) \cite{paysan20093d} is the first widely used 3DMM, which models the shape and texture with the neutral expression. Some later 3DMMs \cite{cao2013facewarehouse, li2017learning} introduce expression bases to model different expression types. For a detailed overview, readers can refer to the survey paper \cite{egger20203d}. 

3DMMs are extensively used for 3D face reconstruction. Many reconstruction methods regress 3DMM coefficients (such as the coefficients of shape and texture) and camera parameters from 2D images. They follow the analysis-by-synthesis paradigm, which minimizes the difference between a rendered image of the 3D reconstructed face and the original image through direct optimization \cite{garrido2016reconstruction, jiang20183d} or deep neural network prediction \cite{deng2019accurate, zhang2021weakly, feng2021learning}.

In this paper, we use the method proposed in \cite{deng2019accurate} to reconstruct the 3D face. This method employs a modified 3DMM based on \cite{paysan20093d} and the expression bases from \cite{cao2013facewarehouse}.


\subsection{Facial Expression Editing}

Facial expression editing refers to the manipulation of facial expressions in images or videos. There are various image-to-image translation methods available, such as \cite{isola2017image, zhu2017unpaired, choi2018stargan}, as well as methods specifically designed for facial expression editing, such as \cite{ding2018exprgan,geng20193d,tewari2020stylerig,tripathy20icface,dapolito21ganmut}). ExprGAN \cite{ding2018exprgan} is a method based on conditional GAN, which can transform faces into specified expressions with continuous intensities. ICface \cite{tripathy20icface} uses head angles and action units to control facial expressions and poses through a two-stage neural network. GANmut \cite{dapolito21ganmut} proposes a GAN-based framework that learns an expressive and interpretable conditional space of emotions. However, these methods do not consider 3D information and face difficulty in disentangling expressions from pose, lip shape, background and other attributes. Geng \textit{et al.} \cite{geng20193d} incorporate a 3DMM in their approach to separate the face into shape and texture, and modify them respectively based on the input 3DMM coefficients, but the quality of the generated image is not satisfactory.

Recently, StyleGAN-based attribute editing has gained attention due to the high editability of StyleGAN and the high quality of the generated results \cite{karras2019style, karras2020analyzing}. This type of attribute editing first projects the input image to StyleGAN's latent space by either optimization \cite{lipton2017precise, karras2020analyzing, abdal2019image2stylegan, abdal2020image2stylegan++}, or regression \cite{richardson2021encoding, tov2021designing, alaluf2021restyle, wang2022high}. Then, the corresponding latent code is moved towards target attribute's location. Finally, StyleGAN generates the edited image from the modified latent code. This approach has advantages over the variational autoencoder (VAE), because VAE inherently tends to generate blurry images \cite{Theis15Evaluation,dumoulin17ALI}.
Nevertheless, finding decoupled editing directions for different facial expressions is laborious and challenging,  because these expressions are found to couple with other attributes such as poses in the latent space of StyleGAN, due to the bias of the training data \cite{shen2020interpreting, harkonen2020ganspace, shen2021closed}.
StyleRig \cite{tewari2020stylerig} incorporates 3DMM into StyleGAN-based editing, where 3DMM coefficients are mapped to StyleGAN's latent code, achieving explicit control over the expression and pose. However, this approach cannot control fine-grained details in textures and may fail if the target expression or pose is absent in the training data. 

It is worth noting that all the methods mentioned above are designed for images; thus, temporal consistency or inter-frame continuity cannot be guaranteed when applying them to video frames. Additionally, most of those methods do not consider lip shape preservation. There are a few methods designed to edit facial expressions in a video. Ma \textit{et al.} \cite{ma2019real} propose a video editing method that reconstructs a 3DMM and transforms the expression coefficients according to the emotion label. In this method, temporal consistency is achieved by window-based smoothing, and lip synchronization is achieved by imposing a lip shape constraint. The Wav2Lip-Emotion method \cite{magnusson2021invertable} extends the lip synchronization architecture \cite{prajwal2020lip} by modifying facial expressions with the aid of $L_1$ reconstruction and pre-trained emotion objectives; however, the face identity in test images is not preserved, and the visual quality of generated images is very low. NED \cite{papantoniou22ned} allows speech-preserving facial expression editing for videos based on 3DMM and neural rendering. However, this method suffers from jitter in the rendered face.
Furthermore, these methods cannot generate facial expressions with different emotional intensities, nor can they seamlessly transition between different emotions. Solanki \textit{et al.} \cite{solanki22deepmanip} propose a method that allows the user to continuously control the intensity of the target emotion when editing a video. However, it cannot preserve the lip shape, making it unsuitable for talking face videos. More recently, Liu \textit{et al.} \cite{liu22DeepFaceVideoEditing} propose a sketch-based facial video editing framework built on StyleGAN3 \cite{karras2021stylegan3}, where they map video frames to the latent space of StyleGAN3 and guide the editing through user-drawn sketches and masks. Their method can generate smooth transitions with good visual quality and temporal consistency. However, when editing talking videos, it is difficult to draw the sketches around the mouth area, as editing may cause the lip out of sync with the audio.

\begin{figure*}[!t]
    \centering
    \includegraphics[width=\textwidth]{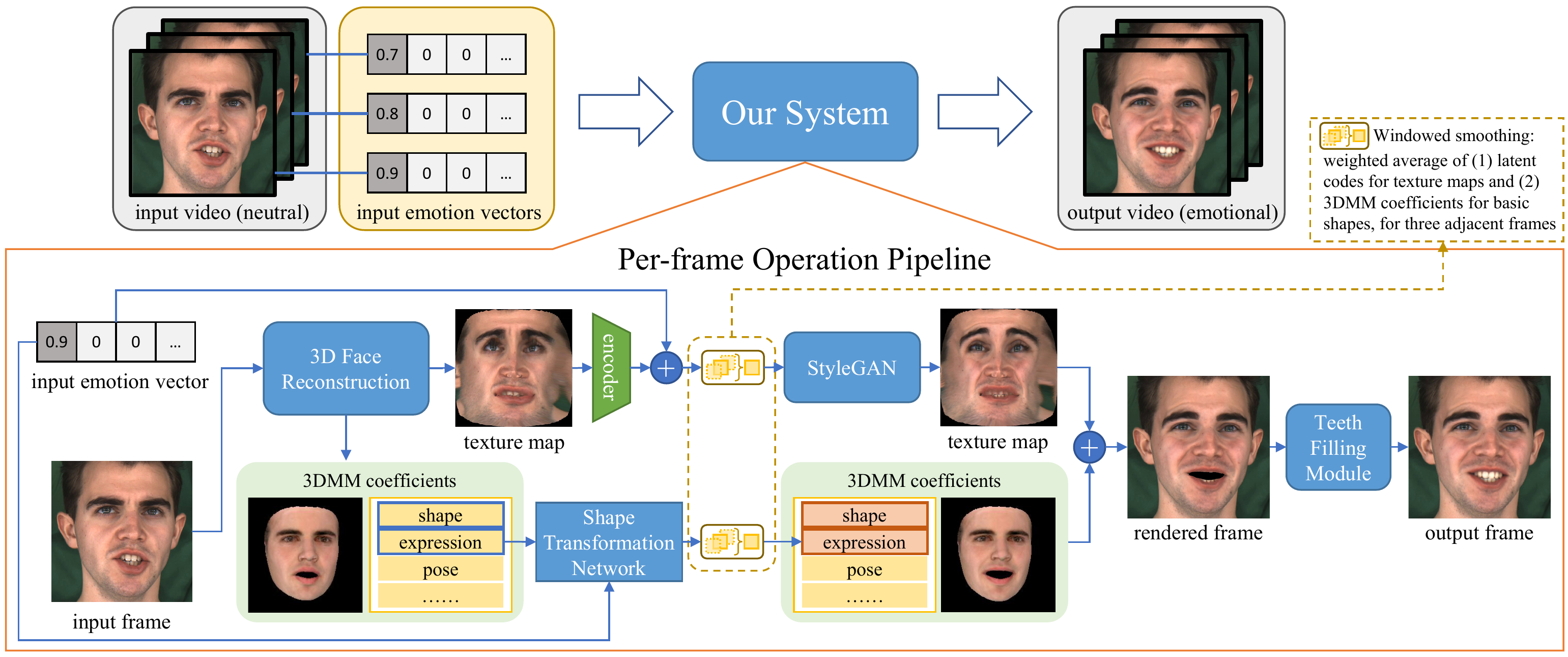}
    \caption{
        The pipeline of our method. (Top row) Facial expression editing is done frame by frame, and for each frame, a windowed smoothing operation is applied before rendering the final output. (Bottom row) Given a video frame, we first reconstruct a 3D face to obtain the 3DMM coefficients and texture map. Then we transform the shape and the texture respectively. For the texture, we use an encoder to map it into a latent code and add an editing direction vector inferred from the emotion vector to generate the transformed texture from the modified latent code by the StyleGAN. For the shape, we use a shape transformation network to transform the 3DMM coefficients according to the emotion vector. Windowed smoothing is applied to ensure inter-frame continuity.  Next, we render the face with the new shape and texture. Finally, we use a teeth filling module to fill the cavity in the teeth area.
    }
    \label{fig:pipeline}
\end{figure*}

\subsection{Audio-driven Talking Face Video Generation}

Audio-driven talking face video generation aims to synthesize realistic-looking talking videos that are synchronized with input audio and have the same facial identity as input face images or video clips \cite{chen2020comprises}. To achieve this goal, motion features need to be predicted from input audio and applied to input visual data. Some methods predict modified pixels directly in the image space \cite{vougioukas2020realistic, prajwal2020lip}, and other methods use the 3DMM \cite{karras2017audio, thies2020neural, yi2020audio} or facial landmarks \cite{chen2019hierarchical} to predict expression coefficients, vertex coordinates, or landmark locations from extracted audio features, which are then used for video generation. Recently, Lahiri \textit{et al.} \cite{lahiri2021lipsync3d} propose explicitly decoupling 3D pose, geometry, texture and lighting in their pipeline, enabling data-efficient learning and high-quality synthesis. However, these methods do not distinguish between speech-related expressions and emotion-related expressions, because it is difficult to infer them simultaneously from the input audio. Therefore, they cannot generate videos with explicit control of facial expressions; in most cases, they can only generate videos that resemble neutral expressions.

A few studies have explored video generation with specified emotions. For example, Karras \textit{et al.} \cite{karras2017audio} include a learned latent representation of emotional states into the input of their system, enabling video generation with different emotions. However, their method requires manual mining for meaningful emotion vectors, which is a demanding task. Moreover, their method only considers shape deformations without modeling textures. Recently, Ji \textit{et al.} \cite{ji2021audio} learn two disentangled latent spaces (a duration-independent emotion space and a duration-dependent content space) using paired audio signals to generate videos with assigned emotions or emotions inferred from the input audio. However, the generated expressions are not accurate enough, and the texture of the face is not consistent in a video, resulting in degraded expressiveness and noticeable artifacts. The method also does not perform well on lip synchronization, probably due to the coupling of expressions and audio signals. 
Ye \textit{et al.}~\cite{ye2022dynamic} use Wav2Lip \cite{prajwal2020lip} to generate the face geometry from the input audio and a dynamic neural texture module to sample the face texture based on the input emotion label. However, their method cannot produce results with high expression intensity properly and sometimes has a color shift problem.

To summarize, generating natural talking face videos with different expressions from the input audio is very challenging. Given that many existing methods can generate good talking face videos with a neutral expression, our work offers a simple yet effective solution to edit facial expressions in talking videos.

\section{Method}

\subsection{Overview}

In this paper, we address the following problem: 

\begin{problem}
Editing the expressions in a talking face video according to prespecified emotion labels (including both emotion type and intensity)) so that the edited video preserves the face identity and maintains audio-lip synchronization.
\end{problem}

As aforementioned, existing image-to-image translation methods and image attribute editing methods cannot solve this problem, because they cannot deal with inter-frame continuity and the coupling of facial expressions and head poses. In this paper, we propose a simple yet effective solution to this problem from a new perspective: noting that when speaking, a person's facial expression and mouth movement are controlled by facial muscles, we address Problem 1 from the perspective of two-level motion information editing. That is, we use a 3DMM to capture major facial movements (level 1) and an associated texture map modeled by StyleGAN to capture appearance fine details and subtle variations (level 2). The overall pipeline of our method is illustrated in Figure \ref{fig:pipeline}.

The input of our method includes a talking face video and an emotion vector for each frame\footnote{If only emotion vectors of some keyframes are provided, we use linear interpolation to assign an emotion vector for each frame.}. Our method outputs an edited video with emotions as specified by the emotion vectors. The emotion vector indicates both expression type and intensity, which is defined as an $n_e$-dimensional non-negative vector $\boldsymbol e = (e_1, \dots, e_{n_e}) \in \mathbb R_{+}^{n_e}$, where $n_e$ is the number of expression types (except for the neutral expression). There is at most one non-zero element in $\boldsymbol e$. Each dimension in $\mathbb R_{+}^{n_e}$ stands for an expression type (such as happiness, sadness, etc.) except for the neutral expression, and the value $e_i$ indicates the intensity of that expression. 
The zero vector represents the neutral expression. Throughout this paper, the intensity is linearly normalized to $[0,1]$.

The two-level motion information editing is performed frame by frame, and for each frame, a windowed smoothing operation is applied for three adjacent frames to ensure inter-frame continuity. In more detail, at a discrete time $t$, a single frame $\boldsymbol I^t$ and a corresponding emotion vector $\boldsymbol e^t$ are input into the system. First, we reconstruct a 3D face (using 3DMM) and extract the texture from $\boldsymbol I^t$ (Section \ref{sec:3d_reconstruction}). Then we use a GAN model to transform the 3DMM coefficients (Section \ref{sec:shape_transformation}) , which is the level-1 modification for capturing major facial movements. We also edit the texture in the latent space of StyleGAN according to the emotion vector $\boldsymbol e^t$ (Section \ref{sec:texture_transformation}), which is the level-2 modification to capture appearance fine details and subtle variations. By combining level-1 and level-2 modifications, we achieve continuous control of the emotion types and intensities. At the end of level-1 and level-2 modifications, we apply a windowed smoothing operation (Section \ref{sec:windowed_smoothing}) to ensure inter-frame continuity. Next, we render the 3D face with the transformed shape and texture onto the original background. Because the 3DMM does not model the teeth area, we use a teeth filling module (Sec. \ref{sec:teeth_completion}) to fill the cavity. Finally, we combine the generated frames together to obtain the edited video.

\begin{figure}[!ht]
\centering
\includegraphics[width=0.48\textwidth]{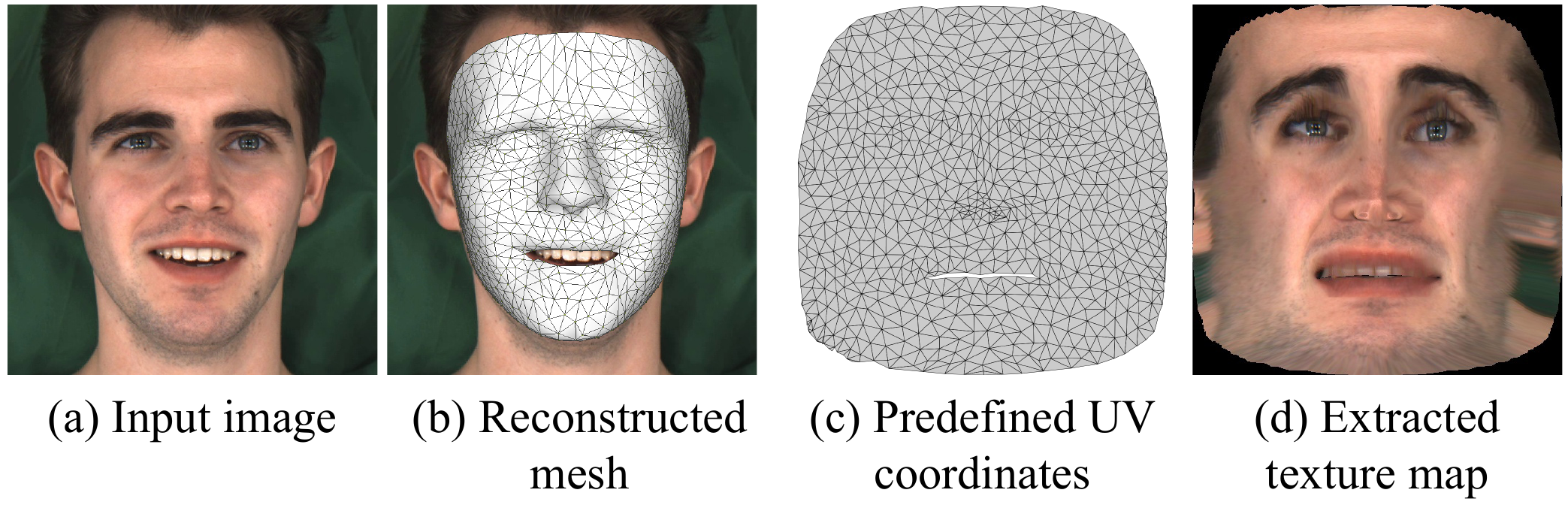}
\caption{We extract the texture from the input image based on the reconstructed mesh and predefined UV coordinates.}
\label{fig:texture_extraction}
\end{figure}

\subsection{3D Face Reconstruction and Texture Extraction} \label{sec:3d_reconstruction}

Given a video frame $\boldsymbol I$ (Fig.~\ref{fig:texture_extraction}a), we reconstruct a 3D face using a state-of-the-art method \cite{deng2019accurate}. This method regresses coefficients of a 3DMM face model from $\boldsymbol I$. The coefficients are in the form of $(\boldsymbol \alpha, \boldsymbol \beta, \boldsymbol \delta, \boldsymbol \gamma, \mathbf p) \in \mathbb R^{257}$, where $\boldsymbol \alpha \in \mathbb R^{80}$ is the shape coefficient, $\boldsymbol \beta \in \mathbb R^{64}$ is the expression coefficient, $\boldsymbol \delta \in \mathbb R^{80}$ is the texture coefficient, $\boldsymbol \gamma \in \mathbb R^{27}$ is the lighting coefficient, and $\boldsymbol p \in \mathbb R^{6}$ is the pose coefficient including rotation and translation. The reconstructed {\it pose-independent} 3D face shape $\mathbf S$ is represented by $\mathbf S = \mathbf {\bar S} + \mathbf B_{shape} \boldsymbol \alpha + \mathbf B_{exp} \boldsymbol \beta$, where the mean shape $\mathbf {\bar S}$ and shape bases $\mathbf B_{shape}$ are from the 3DMM called BFM09 \cite{paysan20093d} and the expression bases $\mathbf B_{exp}$ are from FaceWarehouse \cite{cao2013facewarehouse}.

We use predefined UV coordinates (Fig.~\ref{fig:texture_extraction}c) for each vertex of the 3D face mesh $\mathbf S$, and build a mapping $f$ between the UV coordinates and image pixels by reprojecting $\mathbf S$ onto the image plane (Fig.~\ref{fig:texture_extraction}b) using the standard rasterization pipeline. Finally, we extract the texture (Fig.~\ref{fig:texture_extraction}d) of $\mathbf S$ from $\boldsymbol I$ using the mapping $f$.

\subsection{Level-1 Shape Transformation} \label{sec:shape_transformation}

One of the key challenges in our system is how to transform emotion-related expressions while retaining lip synchronization (which preserves speech-related expressions to some extent). In other words, when editing facial expressions, we must ensure that the transformed mouth shape still matches the pronunciation. One possible way is to collect paired training data of 3DMM coefficients that are aligned to the same phoneme under different facial expressions and then learn the mappings. However, it is laborious and difficult to collect such precise paired data. In our system, we design a shape transformation network trained by unpaired training data with a cycle-consistent training scheme similar to \cite{zhu2017unpaired, choi2018stargan} (Section \ref{sec:shape_network_arch}), which supports pose-independent expression type editing and intensity control (Section \ref{ssubsec:type-intensity-edit}). In addition, we design elaborate loss terms to preserve the mouth shape and personal identity (Section \ref{ssubsec:level-1-loss}).

\subsubsection{Network Architecture} \label{sec:shape_network_arch}

The shape transformation network is a GAN model composed of a generator $G$ and a discriminator $D$. The generator is a fully connected network including an input layer, four hidden layers with LeakyReLU activation and an output layer. The input to $G$ consists of the emotion vector $\boldsymbol e$, the shape and expression coefficients $\boldsymbol \alpha$, $\boldsymbol \beta$ of the 3DMM, so the input layer has $n_c + n_e$ neurons, where $n_c$ is the dimension of the concatenated coefficients $\boldsymbol c = (\boldsymbol \alpha , \boldsymbol \beta)$ and $n_e$ is the number of the emotion types except for the neutral expression. Each hidden layer has $n_h$ neurons whose value is determined empirically in Section \ref{sec:implementation}, and the output layer has $n_c$ neurons. 

The discriminator has a structure similar to $G$, except that the input layer has $n_c$ neurons and the output layer consists of two branches: one branch outputs the estimated probability $D_{rf}(\cdot)$ of whether the input signal to $D$ is real, and the other branch estimates the emotion vector $\boldsymbol e^\prime = D_{reg}(\cdot)$ in the input signal to $D$.

\subsubsection{Expression Type Editing and Intensity Control}
\label{ssubsec:type-intensity-edit}

Given the 3DMM coefficients $\boldsymbol c= (\boldsymbol \alpha , \boldsymbol \beta) \in \mathbb R^{n_c}$ and the target emotion vector $\boldsymbol e \in \mathbb R^{n_e}$, the {\it pose-independent} editing of expression types with intensity control at the level-1 shape transformation is achieved by using the generator $G$ to predict new coefficients $G(\boldsymbol c, \boldsymbol e) \in \mathbb R^{n_c}$. Then the edited 3D shape with the original pose in the input frame $\boldsymbol I$ can be reconstructed by the 3DMM with the coefficients $\boldsymbol c^\prime= (\boldsymbol \alpha^\prime, \boldsymbol \beta^\prime) = G(\boldsymbol c, \boldsymbol e)$ and the pose coefficient $\boldsymbol p$.

\subsubsection{Loss Function}
\label{ssubsec:level-1-loss}

We adopt the Wasserstein GAN objective with gradient penalty \cite{arjovsky2017wasserstein, gulrajani2017improved} to calculate the adversarial loss $\mathcal L_{adv}$ and the gradient penalty loss $\mathcal L_{gp}$, using the output of the probability branch $D_{rf}$ of the discriminator. To ensure that the generated coefficients match the given emotion vector, we propose a regression loss term. For the discriminator,
\begin{equation}
    \mathcal L_{reg}^D = \mathbb E_{\boldsymbol c, \widetilde{\boldsymbol e}} \left\| \widetilde{\boldsymbol e} - D_{reg}(\boldsymbol c) \right\|^2,
\end{equation}
where $\widetilde{\boldsymbol e}$ is the true emotion vector (inferred from the emotion labels in the training data) associated with the shape coefficients $\boldsymbol c$. 
And for the generator,
\begin{equation}
    \mathcal L_{reg}^G = \mathbb E_{\boldsymbol c, \boldsymbol e} \left\| \boldsymbol e - D_{reg}(G(\boldsymbol c, \boldsymbol e)) \right\|^2.
\end{equation}

Denote the 3D mesh shape reconstructed from 3DMM coefficients $\boldsymbol c= (\boldsymbol \alpha , \boldsymbol \beta)$ using the bases $(\mathbf B_{shape}, \mathbf B_{exp})$ by $\mathbf S(\boldsymbol c)$ and its vertex-wise vector form by $\mathbf V_{\mathbf S(\boldsymbol c)}$, i.e., each entity $\mathbf v_i\in\mathbf V_{\mathbf S(\boldsymbol c)}$ is the $(x,y,z)$ coordinate of the $i$th vertex in the mesh $\mathbf S(\boldsymbol c)$.
In our experiments, we find that measuring the difference in the 3D shape space rather than in the coefficient space leads to better results. Thus, we propose the cycle-consistency loss $\mathcal L_{rec}$ as
\begin{equation} \label{eq:L_rec_revised}
\begin{array}{r}
     \mathcal L_{rec} = \mathbb E_{\boldsymbol c, \boldsymbol 0, \boldsymbol e} \left\| \mathbf V_{\mathbf S(\boldsymbol c)}-\mathbf V_{\mathbf S(G(G(\boldsymbol c,\boldsymbol e),\boldsymbol 0)} \right\|_2^2 + \\
     E_{\boldsymbol c^*, \boldsymbol e^*, \boldsymbol 0} \left\| \mathbf V_{\mathbf S(\boldsymbol c^*)}-\mathbf V_{\mathbf S(G(G(\boldsymbol c^*,\boldsymbol 0),\boldsymbol e^*)} \right\|_2^2 
\end{array}
\end{equation}
where $\boldsymbol c$ is a shape coefficient with a neutral expression, $\boldsymbol c^*$ is a shape coefficient with a non-neutral expression $\boldsymbol e^*$, 
both sampled from the dataset. The target emotion vectors are set to a random non-neutral expression $\boldsymbol e$ and a neutral expression $\boldsymbol 0$, respectively.
Note that $G(G(\boldsymbol c,\boldsymbol e),\boldsymbol 0)$ transforms the edited shape $G(\boldsymbol c,\boldsymbol e)$ with expression $\boldsymbol e$ back into a shape with the neutral expression and similar explanation for $G(G(\boldsymbol c^*,\boldsymbol 0),\boldsymbol e^*)$ holds.

We propose a mouth shape preservation loss to force the transformed mouth shape to be similar to the original one. Different from \cite{ma2019real} that calculates the absolute 3D distance of selected key points between the input and transformed lip vertices, we first measure the relative distance from the upper lip to the lower lip and then calculate the difference of the relative distance between the input and transformed one. We believe the relative distance can better model the opening and closing of the mouth and is less affected by lateral movements. This is especially important for conveying emotions such as contempt, where the mouth is often slightly puckered to one side. The loss $\mathcal L_{mouth}$ is proposed as
\begin{equation} 
\label{eq:L_mouth}
\begin{array}{l}
    \mathcal L_{mouth} = \mathbb E_{\boldsymbol c, \boldsymbol e} \|\left(\mathbf V_{\mathbf S_u(\boldsymbol c)} - \mathbf V_{\mathbf S_d(\boldsymbol c)} \right) - \\
    \left(\mathbf V_{\mathbf S_u(G(\boldsymbol c,\boldsymbol e))} - \mathbf V_{\mathbf S_d(G(\boldsymbol c,\boldsymbol e))}\right)\|_2^2 +
    \mathbb E_{\boldsymbol c^*} \|  \left(\mathbf V_{\mathbf S_u(\boldsymbol c^*)} - \mathbf V_{\mathbf S_d(\boldsymbol c^*)} \right) \\
    - \left(\mathbf V_{\mathbf S_u(G(\boldsymbol c^*,\boldsymbol 0))} - \mathbf V_{\mathbf S_d(G(\boldsymbol c^*,\boldsymbol 0))} \right) \|_2^2, 
\end{array}
\end{equation}
where $\mathbf V_{\mathbf S_u(\boldsymbol c)}$ is the vector of selected key points of the upper lip in the mesh corresponding to the 3DMM coefficient $\boldsymbol c$, i.e., each element $\mathbf v_{u_i}\in\mathbf V_{\mathbf S_u(\boldsymbol c)}$ is the $(x,y,z)$ coordinate of the $u_i$th key points of the upper lip. $\mathbf V_{\mathbf S_d(\boldsymbol c)}$ is defined similarly for the lower lip.

Finally, to avoid producing large and abnormal distortion, we add a regularizing term to constrain the deformation caused by the shape coefficient $\boldsymbol \alpha$. We measure the difference of the mesh reconstructed from only the original and generated shape coefficient, i.e., $(\boldsymbol \alpha, \boldsymbol 0)$ and $(\boldsymbol \alpha^\prime, \boldsymbol 0)$ where the $\boldsymbol \alpha$ and $\boldsymbol \alpha^\prime$ are taken from $\boldsymbol c$ and $G(\boldsymbol c,\boldsymbol e)$ and the expression coefficients are both set to $\boldsymbol 0$:
\begin{equation}
    \mathcal L_{r} = \mathbb E_{\boldsymbol c, \boldsymbol e} \left\| \mathbf V_{\mathbf S((\boldsymbol \alpha, \boldsymbol 0))} - \mathbf V_{\mathbf S((\boldsymbol \alpha^\prime, \boldsymbol 0))} \right\|_2^2.
\end{equation}
To summarize, the objective functions for $G$ and $D$ are
\begin{equation}
    \mathcal L_{D} = -\mathcal L_{adv} + \lambda_{gp} \mathcal L_{gp} + \lambda_{reg} \mathcal L_{reg}^D
\end{equation}
\begin{equation}
\label{eq:L_G}
    \mathcal L_{G} = \mathcal L_{adv} + \lambda_{reg} \mathcal L_{reg}^G + \lambda_{rec} \mathcal L_{rec} + \lambda_{mouth} \mathcal L_{mouth} + \lambda_{r} \mathcal L_{r}
\end{equation}
where the weights $\lambda_{gp},\lambda_{reg},\lambda_{rec},\lambda_{mouth},\lambda_{r}$ are specified in Section \ref{sec:implementation}.

\subsection{Level-2 Texture Transformation} \label{sec:texture_transformation}

The texture contains a good amount of emotion-related information, such as fine appearance details and subtle variations between frames, which are expressed through the facial colors and fine texture details that the coarse geometric shape cannot represent (e.g., wrinkles). In our method, we train a StyleGAN (Section~\ref{ssubsec:stylegan}) to generate texture maps from edited latent codes, where the edited latent codes are obtained by projecting the input texture maps to the StyleGAN's latent space using an encoder (Section~\ref{ssubsec:encoder}), and then modified based on the emotion vectors (Section~\ref{ssubsec:latent_editing}).

\subsubsection{StyleGAN} \label{ssubsec:stylegan}
The StyleGAN \cite{karras2019style,karras2020analyzing} is a state-of-the-art image generation model introduced by Karras \textit{et al.}, which uses a style-based approach to generate highly realistic images. Given a latent code $\mathbf z$ in the input latent space $\mathcal Z$, a mapping network $f: \mathcal Z \rightarrow \mathcal W$ first map it to an intermediate latent code $\mathbf w$ in the intermediate latent space $\mathcal W$. Then, $\mathbf w$ is replicated and converted by $n$ learned affine transformations $\{A_i | i=1, 2, \dots, n\}$ into $n$ styles, which are used to modulate the $n$ convolution layers of StyleGAN's synthesis network. The synthesis network generates the output image by upsampling from a learned constant input, guided by the styles. We denotes this image generation process using the following equations, where $\hat t$ is the generated image:
\begin{equation}
    \mathbf w = f(\mathbf z),
\end{equation}
\begin{equation}
    \hat t = \mathrm{stylegan}\left( A_1(\mathbf w), A_2(\mathbf w), \dots, A_n(\mathbf w) \right).
\end{equation}

Specifically, for each actor, we train a StyleGAN-based model \cite{karras2020analyzing} using the texture maps extracted from the training videos. More training details can be found in Section~\ref{sec:implementation}.

\subsubsection{Encoder} \label{ssubsec:encoder}

Given a real image $t$, StyleGAN itself cannot determine the corresponding latent code $\mathbf z$ or the intermediate latent code $\mathbf w$ for the most accurate image reconstruction. Moreover, the $\mathcal Z$ or $\mathcal W$ space has been found to have limited capacity to express real images. Therefore, it is common practice to invert images into an extended latent space $\mathcal W+$, defined by the stacking of $n$ different $\mathbf w$ codes, by either optimization \cite{lipton2017precise, karras2020analyzing, abdal2019image2stylegan, abdal2020image2stylegan++}, or regression \cite{richardson2021encoding, tov2021designing, alaluf2021restyle}. The process of inversion and reconstruction can be expressed as follows:
\begin{equation} \label{eq:stylegan_inversion}
    \mathbf{W} := \{\mathbf w_1, \mathbf w_2, \dots, \mathbf w_n\} = \mathrm{inv}(t),
\end{equation}
\begin{equation} \label{eq:stylegan_w+}
    \hat t = \mathrm{stylegan}\left( A_1(\mathbf w_1), A_2(\mathbf w_2), \dots, A_n(\mathbf w_n) \right),
\end{equation}
where the reconstructed image $\hat t$ is expected to be similar to the real image $t$. For simplicity, we also express Eq.(\ref{eq:stylegan_w+}) as $\hat t = \mathrm{stylegan}(\mathbf{W})$ with a slight abuse of notation.

As optimization is time-consuming, we adopt an encoder \cite{richardson2021encoding} which maps the real image $t$ into the $\mathcal W+$ space by regression. The encoder first extracts multi-resolutional feature maps from $t$ using a standard feature pyramid over a ResNet backbone \cite{deng2019arcface}. It then applies $n$ fully convolutional networks to transform the feature maps into $n$ $\mathbf w$ codes. We represent this process as $\mathbf{W} = \mathrm{enc}(t)$, which is analogous to Eq.(\ref{eq:stylegan_inversion}).

We use the trained StyleGAN in the previous section as the decoder with fixed weights and train the encoder by minimizing the pixel difference and perceptual distance between the input training texture map $t$ and reconstructed texture map $\hat t = \mathrm{stylegan}(\mathrm{enc}(t))$. Refer to Section~\ref{sec:implementation} for further details.

\subsubsection{Texture Editing in the Latent Space} \label{ssubsec:latent_editing}

After training the StyleGAN and encoder for the texture maps, we calculate the editing direction vector in the StyleGAN's latent space for each non-neutral emotion. Denote the set of all the training texture maps with the highest intensity of emotion $x$ as $T_x$ (for the neutral emotion $x'$, all the texture maps of $x'$ are selected into $T_{x'}$), we compute the editing direction vectors in the following two steps: (1) for each emotion $x$, map all the texture maps in $T_x$ to the latent space and calculate the average latent code, and (2) for each non-neutral emotion $y$, calculate the difference between the average latent code of $y$ and the neutral emotion, which yields the editing direction vector $d_y$. These two steps can be formulated as:
\begin{equation}
d_{y}=\mathbb E_{t_1\in T_{y}} \left(\mathrm{enc}(t_1) \right) - \mathbb E_{t_2\in T_{neutral}}\left(\mathrm{enc}(t_2) \right).
\end{equation}

During editing, given an extracted texture $t_{\boldsymbol 0}$ with the neutral emotion and an emotion vector $\boldsymbol e$ with only one non-zero element $e_y$ (which indicates the target emotion type is $y$ and the intensity is $e_y$), the texture transformation is performed by:
\begin{equation}
t_{\boldsymbol e} = \mathrm{stylegan}(\mathrm{enc}(t_{0}) + e_y \cdot d_y),
\end{equation}
where we first get the latent code of $t_{\boldsymbol 0}$, add the intensity-weighted editing direction vector, and then let the StyleGAN generate the transformed texture from the modified latent code.

Note that in order to make the intensity change of transformed texture as linearly as possible, we only use the textures with the highest emotion intensity to calculate the editing direction vectors, rather than considering all intensities and calculating the editing direction vectors for each intensity range. The reason is that the former makes the modified latent code a linear function of the intensity, while the latter may lead to different rates of change of the latent code in different intensity ranges. Furthermore, since the texture captures the appearance details, a texture that varies linearly with the intensity will greatly contribute to the final rendered face expression that varies linearly with intensity.

\subsection{Refinement}
After transforming the shape (level-1) and texture (level-2), we render the new face with modified expressions and paste it back into the original image. Several special treatments need to be considered to achieve high-quality expression editing, lip synchronization and inter-frame continuity, which are presented in the following subsections.

\subsubsection{Windowed Smoothing} \label{sec:windowed_smoothing}

If smoothing is not applied, the generated shapes and textures may be temporally discontinuous (e.g., containing jitters) because they are obtained in a frame-by-frame manner. Thanks to our decoupled representations of the shape and texture, we can apply a simple operation that smooths the 3DMM coefficients in the shape space and the latent codes in the texture space. I.e., for the frame $\boldsymbol I^t$, we average the edited 3DMM coefficients and latent codes in $\boldsymbol I^{t-1}$, $\boldsymbol I^t$ and $\boldsymbol I^{t+1}$ using the weight of a Hanning window.

Benefiting from our elaborately designed two-level expression representation, the simple smoothing operation works, because (1) the 3DMM we adopt is a linear model, and (2) the latent codes in the texture space generated from consecutive frames are close to each other (i.e., they can be thought of as lying in a hyperplane locally approximating the tangent plane of an expression manifold) and then can be smoothed by a simple average.

\subsubsection{Teeth Completion} \label{sec:teeth_completion}
Note that the 3DMM we adopt does not model the teeth and the interior of the mouth, which leaves a blank area within the mouth in the rendered image, as shown in Figure \ref{fig:pipeline}. Thus, we propose a teeth filling module (see bottom right in Figure \ref{fig:pipeline}) to fill this cavity, which is a typical inpainting problem. Since humans are particularly sensitive to the artifacts around the mouth, we utilize a StyleGAN and an encoder which are the same kinds of networks (with smaller sizes) in Section~\ref{sec:texture_transformation}) in this module to ensure high visual quality and inter-frame continuity.

\begin{figure}[!ht]
\centering
\includegraphics[width=0.35\textwidth]{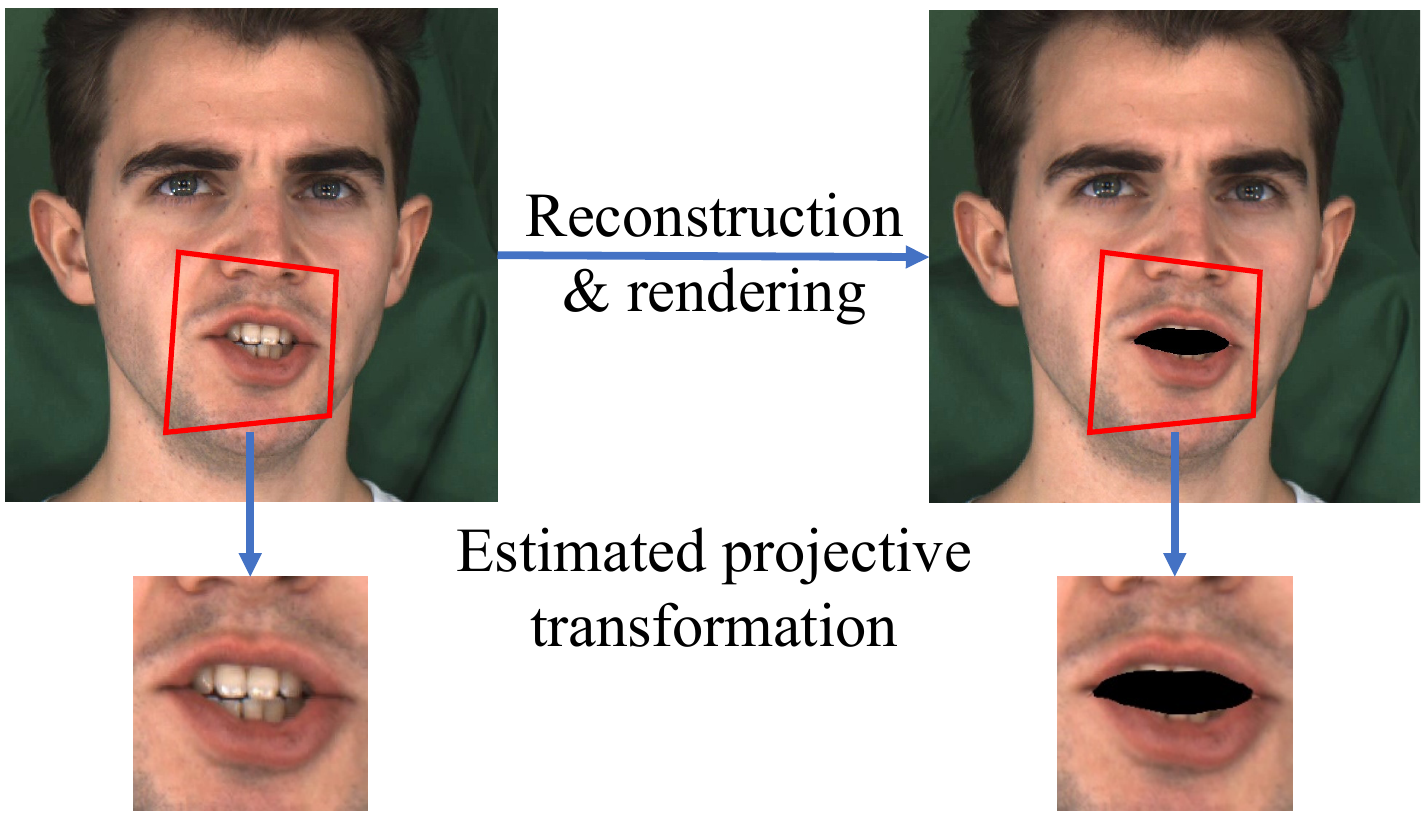}
\caption{We collect training data for the teeth filling module from the original and rendered images.}
\label{fig:mouth_pair}
\end{figure}

We use the following steps to collect the training data for the teeth filling module. First, for all the training frames of an actor, we only run the reconstruction step and the rendering step in our pipeline, i.e., skipping the editing step and teeth filling step, to get paired face images with and without teeth (see the first row in Figure \ref{fig:mouth_pair} for an example). Second, to handle faces with various poses, we estimate a projective transformation to align the mouth area. Specifically, after reconstructing the original image, we first get the 2D mouth landmarks $l$. Then we set the pose parameter of the face model to $\boldsymbol 0$ and project it to the image plane to obtain the frontalized mouth landmarks $l^\prime$. Finally, we estimate the prospective transformation from $l$ and $l^\prime$, apply it to the face images and crop the mouth area  (see the second row of Figure \ref{fig:mouth_pair} for an example).

The mouth images with teeth are used to train a StyleGAN (called the mouth StyleGAN) for producing high-quality mouth images, and the paired mouth images (with and without teeth) are used to train the StyleGAN encoder. Different from the texture encoder, which aims to invert input textures into a latent code for later reconstruction by the texture StyleGAN, the mouth encoder functions more like a translation network. It learns to map the toothless mouth image to a latent code, which the mouth StyleGAN can then use to recover the paired toothed image. The training parameters can be found in Section~\ref{sec:implementation}.

During editing, given a rendered frame without teeth, we align and crop the mouth area in the same way, map it to a toothed latent code with the encoder, and use the mouth StyleGAN to generate the mouth image with teeth from the latent code. Then, we paste it back to the rendered frame to obtain the final output frame.

\subsubsection{Masked Blending}
To seamlessly blend the rendered face with the original background, we use a soft mask where the value gradually increases from 0 to 1 near the edge of the face. We get this mask by first eroding the original hard mask and then applying a Gaussian blur step.

\section{Experiment}

\subsection{Experiment Setup}

\subsubsection{Datasets}

We conduct experiments on two datasets: MEAD \cite{wang2020mead} and RAVDESS \cite{livingstone18ravdess}. The MEAD dataset contains high-quality talking-face videos of 60 actors with eight categories of emotions (i.e., neutral, happy, sad, angry, fearful, surprised, disgusted and contemptuous) at three intensity levels, recorded from seven different view angles. For every actor, there are about 30 videos for each intensity of each emotion at each view angle. We select four actors and for each of them, we sample one out of every five frames from the videos with the frontal view to obtain experimental data, resulting in about 15,000 frames per actor. To show the generalization of our model, we also conduct experiments on RAVDESS \cite{livingstone18ravdess} (Section \ref{sec:additional_results}), another emotional audio-visual dataset with eight categories of emotions (i.e., neutral, happy, sad, angry, fearful, surprised, disgusted and calm) at two intensity levels. However, each intensity has only four frontal-view videos, which are far fewer than MEAD. Therefore, we extract all the video frames (about 7,000 frames per actor) for the experiment.

\subsubsection{Implementation and Training Details}
\label{sec:implementation}
Our method is implemented using PyTorch. The parameters for the shape transformation network are set as follows: $n_c = 144$, $n_e = 7$, $n_h=512$ and $\lambda_{gp} = 10$, $\lambda_{reg} = 20$, $\lambda_{rec} = 5e^3$, $\lambda_{mouth} = 1$, $\lambda_r = 1e^3$. For each actor, we train the shape transformation network using Adam optimizer with $lr = 1e^{-3}$, $\beta_1$ = 0.5 and $\beta_2$ = 0.999 for 80k iterations with a batch size of 64.

For each actor, we train a $512 \times 512$ StyleGAN for the texture with a batch size of 16 for 100k iterations, and train its encoder with a batch size of 4 for 100k iterations. For the teeth filling module, we train a $128\times 128$ StyleGAN with a batch size of 32 for 30k iterations, and train its encoder with a batch size of 4 for 60k iterations. The other settings of the StyleGAN and the encoder are the same as in \cite{karras2020analyzing} and \cite{richardson2021encoding}.

Benefiting from the powerful capabilities of StyleGAN, our method is easily generalized to produce results of higher resolution.
To show this capacity, we also train a model with a $1024 \times 1024$ texture StyleGAN and a $256 \times 256$ teeth StyleGAN. This high-resolution model can produce much finer grained details. The corresponding results can be found in the supplemental demo video.

\subsection{Evaluation Metrics}
\label{subsec:metrics}
We use the following metrics to evaluate the effect of expression editing, lip synchronization and image quality of the generated results.

\textbf{Linearity of intensity editing (LIE).} To easily control the editing of expression intensity using a scale value $e_y$ in $[0,1]$, it is desired that the change in perceived expression intensity is linear with the change in $e_y$. To evaluate this linear property, we design a new perception-based scale-invariant metric called LIE. Given an image $x$ with the neutral expression and a target emotion type $y$, we first generate $n+1$ images from $x$ using emotion vectors of type $y$ whose intensities increase linearly from 0 to 1. Then we use the LPIPS \cite{zhang2018unreasonable} model to calculate the pairwise perceptual difference between neighboring images (i.e., having neighboring intensity values), denoted as $d_i$ ($i=1,2,\dots,n$). Finally, we compute the LIE, which is defined as the coefficient of variation of $d_i$:
\begin{equation}
    \text{LIE} = CV(d_i) = \frac{\sigma(d_{i})}{\mu(d_{i}) },
\end{equation}
where $\sigma(d_{i})$ is the standard deviation of $d_{i}$ and $\mu(d_{i})$ is the mean of $d_{i}$. Note that LIE is non-negative, and when the map from $e_y$ to the perceived expression intensity is linear, LIE is exactly zero.

\textbf{Fr\'echet emotion/inception distance (FED/FID).} Apart from the Fr\'echet inception distance (FID) \cite{heusel2017gans} that has been widely used to assess the quality of images created by a generative model, we use the Fr\'echet emotion distance (FED), where the Inception model in FID is replaced with an emotion recognition network \cite{savchenko2021facial} trained on the AffectNet \cite{mollahosseini2017affectnet} dataset which has the same emotion categories as in MEAD, to assess the effect (e.g., correctness and similarity) of expression editing.

\textbf{Other metrics.} We use the cumulative probability of blur detection (CPBD) \cite{narvekar2011no} to assess the sharpness of generated faces, and cosine similarity of the average ArcFace \cite{deng2019arcface} feature for identity preservation (ID). We also use LSE-D and LSE-C from SyncNet \cite{prajwal2020lip, chung2016out} to assess the lip synchronization of generated videos.

\subsection{Ablation Study}

\begin{figure}[!ht]
\centering
\includegraphics[width=0.35\textwidth]{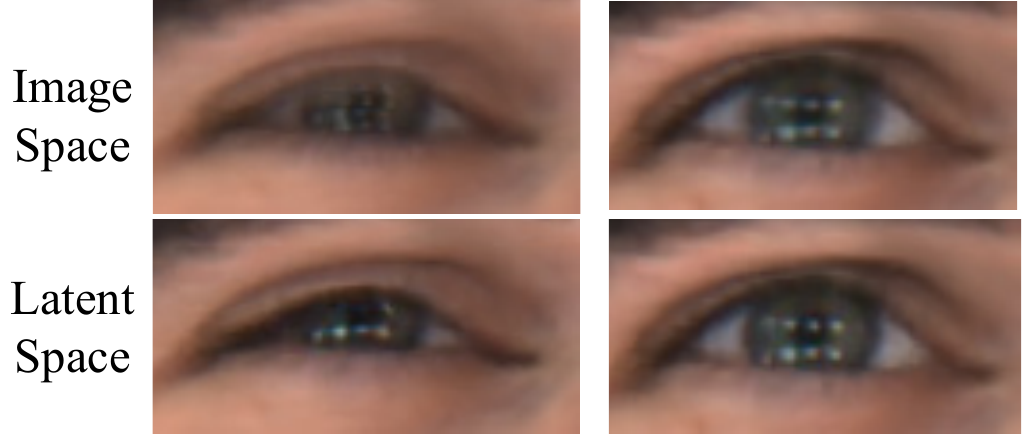}
\caption{Comparison of smoothing in image space and latent space.}
\label{fig:texture_smoothing}
\end{figure}

\subsubsection{3DMM-based Face Representation}
As noted in previous sections, while StyleGAN is capable of generating high-quality facial images and has been used for expression editing, preserving attributes like pose while only modifying the expression during editing is challenging, especially when strong biases exist in the relationship between expressions and poses. Therefore, we employ a 3DMM to first convert the face into a pose-independent shape and texture before editing each separately. To validate the necessity of such decomposition, we remove all 3DMM-related modules (the face reconstruction module and the shape transformation network) and the teeth filling module. The ablation method, ``ours w/o 3DMM'', trains a StyleGAN and an encoder using face images directly and edits expressions in the latent space of the StyleGAN, which resembles typical StyleGAN-based latent code editing methods \cite{richardson2021encoding, tov2021designing, wang2022high}. Experiments show that the StyleGAN struggles to preserve the actor's pose during editing, as illustrated in the last column of Fig.~\ref{fig:comparison} and the demo video.

\subsubsection{Level-1 and Level-2 Transformation}

Our method uses a two-level facial expression editing strategy, i.e., level-1 shape transformation and level-2 texture transformation. Here we present a study to verify that both levels of transformation are necessary. First, we generate 28 videos of seven non-neutral expression types for four actors using level-1 shape transformation only.

Happiness and surprise are classified as positive expressions, while the remaining are classified as negative expressions. 
We then invite 11 users (seven males and four females, average age 25 years and SD = 2.75 years) and ask them to judge whether the edited expression is positive or negative. The results show that the accuracy is 74\%, indicating that level-1 shape transformation can model the major facial movement and thus is necessary. 

Next, we show that both level-1 and level-2 transformations are necessary. For each actor, we generate 105 videos with seven non-neutral expression types and three intensities from five neutral videos with or without level-1 or level-2 transformation, respectively. Based on the generated videos, we compute the values of FED, FID and ID metrics, and the results are summarized in Table \ref{tab:ablation_branch}. The results show that compared with only using level-1 transformation, using both levels of transformations can significantly improve the performance on three metrics, indicating that level-2 texture transformation can capture appearance details and their subtle changes between adjacent frames and thus is also necessary.

\begin{table}[!ht]
\centering
    \renewcommand{\arraystretch}{1.3}
    \caption{Ablation study \textit{w.r.t.} of level-1 and level-2 transformation.}
    \label{tab:ablation_branch}
    \centering
    \begin{tabular}{c|ccc}
    \hline
    Metric          & full           & w/o lvl-1 & w/o lvl-2 \\
    \hline
    FED$\downarrow$ & \textbf{9.40} & 14.32    & 22.82       \\
    FID$\downarrow$ & \textbf{30.36} & 32.48     & 37.76     \\
    ID$\uparrow$    & \textbf{0.931} & 0.900    & 0.849      \\
    \hline
    \end{tabular}
\end{table}

\subsubsection{Smoothing Operation}

To verify the effectiveness of the smoothing operation, we generate 14 pairs of videos with or without this operation (other conditions are the same). We invite eight users (four males and four females, average age 26 years and SD = 2.52 years), and asked them to choose in each pair the video that is better in terms of inter-frame continuity, audio-visual synchronization and visual clarity. The results (Table \ref{tab:ablation_smooth}) show that our proposed smoothing operation significantly improves the inter-frame continuity and slightly improves visual clarity (by mitigating jitters). Interestingly, we find that the smoothing operation also helps to improve audio-visual synchronization, probably because they can alleviate some abrupt changes caused by noise during generation.

\begin{table}[!ht]
\centering
    \renewcommand{\arraystretch}{1.3}
    \caption{The user study of the proposed smoothing operations. The users are required to choose the one that performs better or choose the "same" if both perform similarly. }
    \label{tab:ablation_smooth}
    \centering
    \begin{tabular}{c|ccc}
    \hline
    Preference & Smoothed & Unsmoothed & Same \\ \hline
    continuity & 60\%     & 14\%       & 26\% \\
    sync       & 33\%     & 10\%        & 57\% \\
    clarity    & 19\%     & 12\%       & 70\% \\ \hline
    \end{tabular}
\end{table}

\begin{figure*}[!ht]
    \centering
    \includegraphics[width=\textwidth]{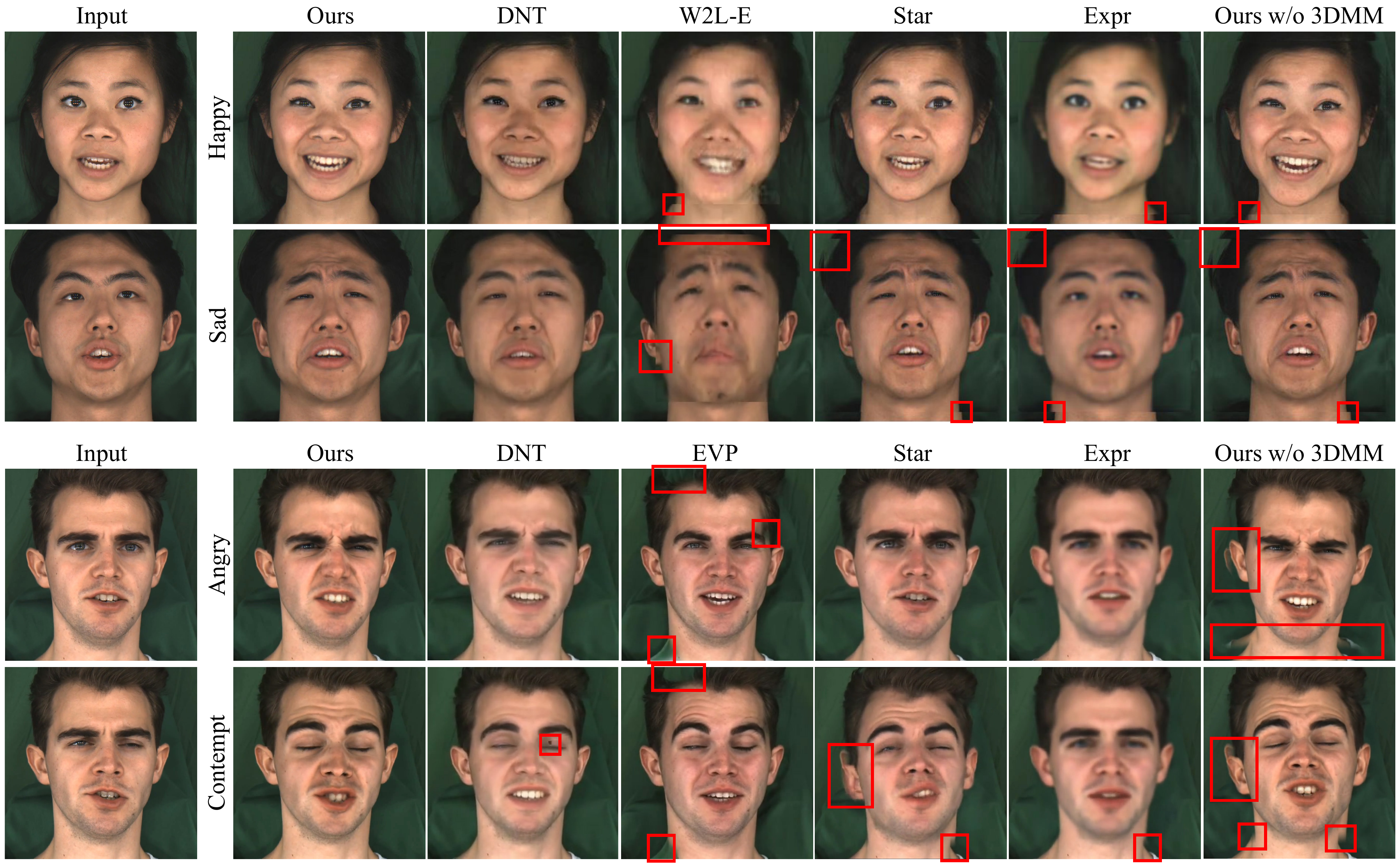}
    \caption{
        Comparison of edited results by different methods. For brevity, we refer to Dynamic Neural Texture \cite{ye2022dynamic} as DNT, Emotional Video Portraits \cite{ji2021audio} as EVP, StarGAN \cite{choi2018stargan} as Star, ExprGAN \cite{ding2018exprgan} as Expr, and Wav2Lip-Emotion \cite{magnusson2021invertable} as W2L-E. The input of all methods contains a video with a neutral emotion and a target emotion vector, and we show the frames from the generated videos. The red boxes indicate some artifacts (e.g., misaligned and eroded areas).
    }
    \label{fig:comparison}
\end{figure*}

\begin{figure}[!ht]
\centering
\includegraphics[width=0.48\textwidth]{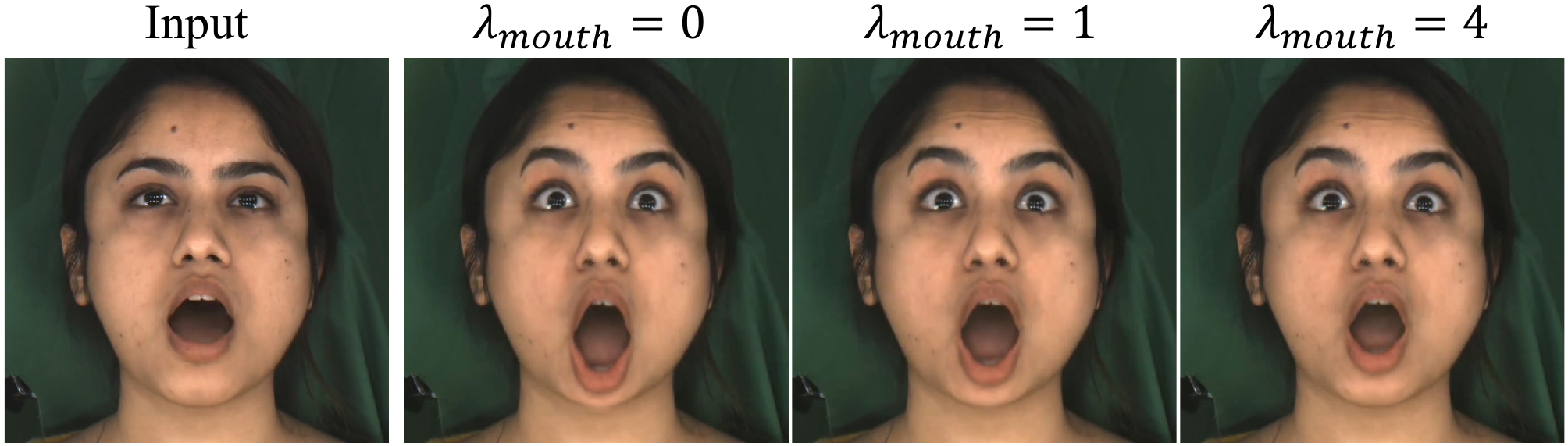}
\caption{The generated expression using different values of $\lambda_{mouth}$.}
\label{fig:ablation_lip}
\end{figure}

Our smoothing operation smooths texture maps using the latent codes in the texture space. To verify this special design, we compare the results of image-space smoothing (i.e., average by pixel information) and latent-space smoothing. Some qualitative results are shown in Figure \ref{fig:texture_smoothing}, from which we observe that the image-space smoothing may lead to blurry or even unrealistic results. In contrast, the latent-space smoothing can consistently produce clear results.

\subsubsection{Mouth Shape Preservation Loss}

We propose a novel mouth shape preservation loss (ref. Eq.(\ref{eq:L_mouth})) in the loss function (Eq.(\ref{eq:L_G})). The weight $\lambda_{mouth}$ of $\mathcal L_{mouth}$ in Eq.(\ref{eq:L_G}) controls the trade-off between the preservation of the original mouth shape and the degree of exaggeration of the edited expression. Some qualitative results are shown in Figure \ref{fig:ablation_lip}, showing that zero or small weights may lead to degraded lip synchronization and abnormal expressions, while large weights may lead to degraded expressiveness of expression editing. We also include quantitative results in Table~\ref{tab:ablation_lip}, where we compare the mouth shape preservation loss based on relative distance, absolute distance, and no mouth shape preservation loss. The results clearly demonstrate that using relative distance yields the best outcome.

\begin{table}[ht]
    \renewcommand{\arraystretch}{1.3}
    \caption{Ablation study \textit{w.r.t.} the mouth shape preservation loss. RD: use relative distance; AD: use absolute distance.}
    \label{tab:ablation_lip}
    \centering
    \begin{tabular}{c|ccc}
    \hline
    Metric            & Ours (RD)     & Ours (AD)   & w/o $\mathcal L_{mouth}$ \\
    \hline
    LSE-D$\downarrow$ & \textbf{8.03} & 8.13      & 8.17         \\
    LSE-C$\uparrow$   & \textbf{7.11} & 7.02       & 6.97       \\
    \hline
    \end{tabular}
\end{table}

\subsection{Comparison with State of the Arts}

We compare our method with three state-of-the-art methods, Dynamic Neural Texture \cite{ye2022dynamic}, Emotional Video Portraits \cite{ji2021audio} and Wav2Lip-Emotion \cite{magnusson2021invertable}, which are designed for generating or editing talking videos with controllable expressions. Since our expression editing task can be addressed by image-to-image translation (I2I) methods, we also compare two state-of-the-art I2I methods, including StarGAN\cite{choi2018stargan} and ExprGAN\cite{ding2018exprgan}. Both methods are conditional-GANs, where StarGAN is a multi-domain I2I method and ExprGAN is dedicated to expression editing for face images. We adapt StarGAN's input and objectives to accept emotion vectors as input. All the methods are  individually trained using cropped face images from the MEAD dataset.

\begin{figure*}[!ht]
    \centering
    \includegraphics[width=\textwidth]{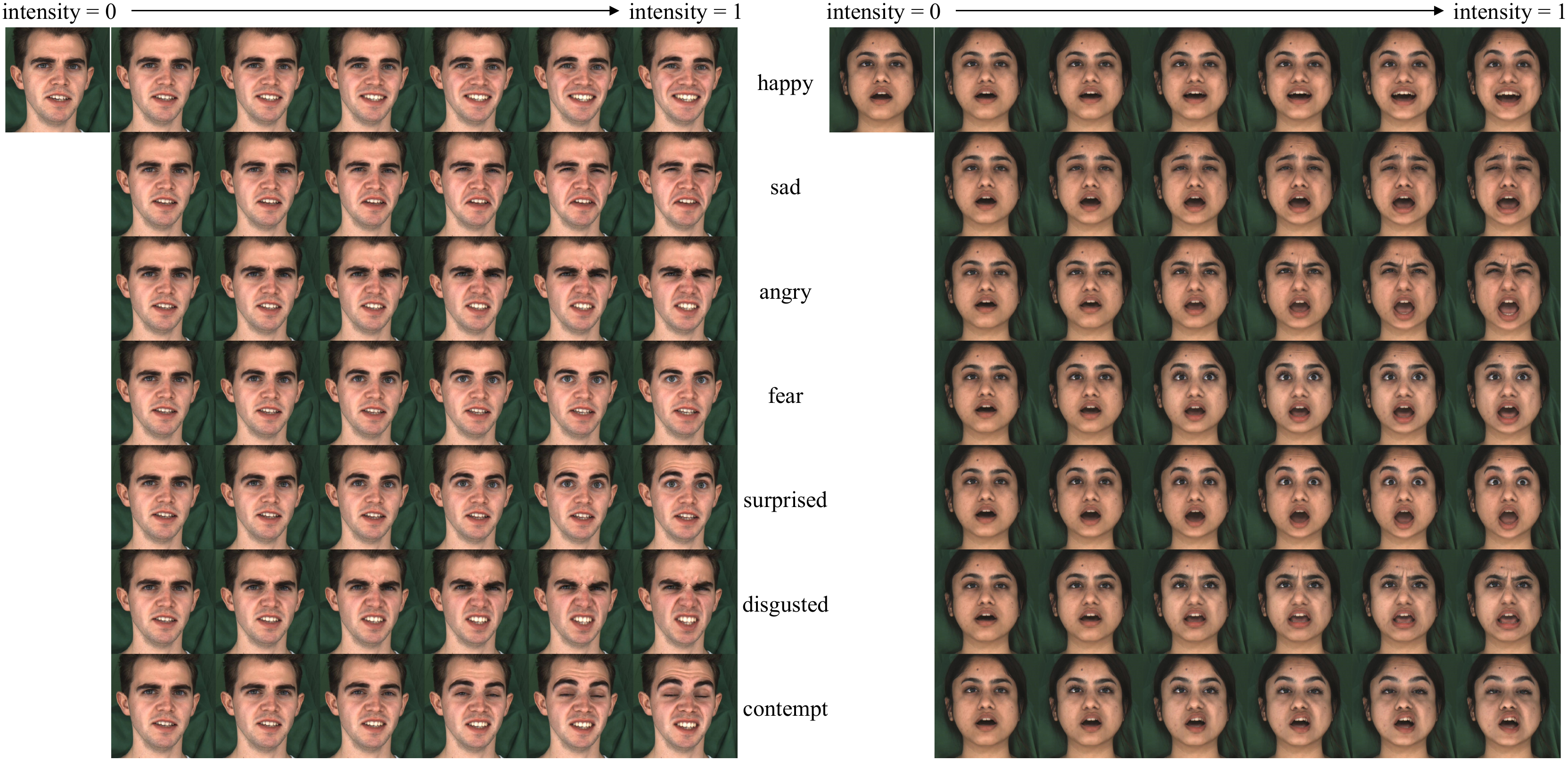}
    \caption{
        Editing results of different facial expression types and intensities generated by our method. The input includes an image with a neutral emotion and the intensity of the target expression type (increasing linearly from 0 to 1). These results show that the edited expressions vary smoothly with different intensity values.
    }
    \label{fig:all_emotions}
\end{figure*}

\subsubsection{Qualitative Evaluation}

We present a qualitative analysis in this section. Some editing results of different identities (male vs. female, white vs. yellow) and expression types (happy, sad, angry and contempt) are illustrated in Figure \ref{fig:comparison} and the supplementary demo video. The input to each method includes a video clip with a neutral expression and a target emotion vector (we also input the original audio clip for DNT and EVP as they require audio input), and the output is a video with the expression type and intensity as specified in the emotion vector.

The visual results in Figure \ref{fig:comparison} and demo video show that I2I methods fail to decouple the expression with the pose, background and other unrelated attributes, resulting in face misalignment and posture changes. The inter-frame continuity of the videos generated by I2I methods is also low. The results of Wav2Lip-Emotion suffer severely from facial distortion and inter-frame discontinuity. DNT and EVP can generate talking videos with desirable facial expressions, but there are still the following problems. The intensity of the generated expressions is not strong enough and the video quality of DNT is not good enough.
EVP produces better video quality, but throughout the video, the texture of the face area is inconsistent (e.g., with teeth and wrinkles moving) and the background gradually erodes the face. This is probably because EVP generates images based on edges, causing the distortion of texture information in edge areas. Moreover, compared to the ground truth in the dataset, some expressions generated by EVP are not accurate. By contrast, our method has good visual quality, high accuracy of expression editing, and good temporal consistency.

Our method is specially designed so that the edited facial expressions can vary smoothly with different intensity values. To demonstrate this property, results with different expression types and intensities are generated by our method, which are shown in Figure \ref{fig:all_emotions}. In particular, our method takes an image with neutral emotion as input, and the intensity values of each expression type increase linearly from 0 to 1. The results show that our method can smoothly change facial expressions (for different types or intensity values) while preserving other attributes. Note that in the training dataset from MEAD, there are only three discrete intensity levels, whereas our method can generate results with intensity values continuously varying in $[0,1]$ by interpolation in the latent space characterizing facial expressions. The reader is also referred to the supplementary demo video for more results and comparisons with other methods.

\begin{table}[!ht]
\renewcommand{\arraystretch}{1.3}
\caption{Quantitative evaluation of average values of the LIE metric and sharpness of edited images measured by the CPBD metric. The average CPBD score of the sampled input images is provided for reference.}
\label{tab:image}
\centering
\begin{tabular}{c|cccc|c}
\hline
Metric          & Ours           & DNT & Star  & Expr  & Input \\
\hline
LIE$\downarrow$  & 0.739 & \textbf{0.582} & 1.320 & 3.317 & N/A   \\
CPBD$\uparrow$  & \textbf{0.247} & 0.225 & 0.193 & 0.102 & 0.280 \\
\hline
\end{tabular}
\end{table}

\begin{table*}[!ht]
\begin{minipage}{0.55\linewidth}
    \renewcommand{\arraystretch}{1.3}
    \caption{Quantitative evaluation of expression editing effect, lip synchronization and image quality on video editing. The average LSE-D, LSE-C and CPBD scores of the input videos are provided for reference.}
    \label{tab:video}
    \centering
    \begin{tabular}{c|cccc|cc|c}
    \hline
    Metric            & Ours           & DNT   & EVP   & W2L-E  & Star          & Expr  & Input \\
    \hline
    FED$\downarrow$   & \textbf{9.40}  & 11.06 & 16.71 & 26.75  & 14.33   & 19.44 & N/A   \\
    FID$\downarrow$   & \textbf{30.36} & 56.42 & 53.42 & 111.65 & 47.47    & 90.19 & N/A   \\
    ID$\uparrow$      & \textbf{0.931} & 0.921 & 0.791 & 0.686  & 0.921    & 0.917& N/A   \\
    \hline
    LSE-D$\downarrow$ & 8.03           & 10.72 & 10.61 & 8.43   & \textbf{7.64}& 8.17 & 7.45  \\
    LSE-C$\uparrow$   & 7.11          & 3.68  & 4.14  & 6.40   & \textbf{7.59} & 7.14 & 7.84 \\
    \hline
    CPBD$\uparrow$    & \textbf{0.166} & 0.121 & 0.152 & 0.124  & 0.096    & 0.024 & 0.213 \\
    \hline
    \end{tabular}
\end{minipage}%
\begin{minipage}{0.45\linewidth}
    \renewcommand{\arraystretch}{1.3}
    \caption{User study results (Acc. for accuracy and cont. for continuity). The users are asked to rate from 1 (lowest) to 5 (highest).}
    \label{tab:user_study}
    \centering
    \begin{tabular}{c|ccccc}
    \hline
    Method & Acc. & Identity & Sync & Cont. & Overall \\
    \hline
    Ours  & \textbf{3.94} & \textbf{4.44} & \textbf{4.38} & \textbf{4.27} & \textbf{4.25} \\
    DNT   & 2.80 & 3.41 & 2.51 & 2.83 & 2.63 \\
    EVP   & 3.11 & 3.72 & 3.20 & 3.56 & 3.32 \\
    W2L-E & 1.78 & 1.38 & 2.16 & 1.20 & 1.19 \\
    Star  & 3.15 & 3.59 & 3.79 & 2.41 & 2.96 \\
    Expr  & 1.76 & 2.57 & 3.29 & 2.07 & 2.06 \\
    \hline
    \end{tabular}
\end{minipage}%
\end{table*}

\subsubsection{Quantitative Evaluation}

First, we evaluate the properties of linear editability and image sharpness measured by the metrics LIE and CPBD based on sampled images. Since Wav2Lip-Emotion does not support specifying the intensity of the emotion, and the code of EVP relevant to this experiment is not available, these two methods are not compared. We sample 100 images with the neutral expression for each actor, and for each expression type, we linearly increase the intensity values from 0 to 1 and generate the edited images: 100 discrete intensity values are used for LIE and 10 values for CPBD. All generated images are resized to $256\times 256$. The results are summarized in Table \ref{tab:image}. Our method achieves the best CPBD score and the second best LIE score among all the methods. 

Next, we evaluate all the methods using metrics FED, FID, ID, LSE-D, LSE-C and CPBD based on the generated videos. Given videos with a neutral expression, we edit them into videos with non-neutral expressions in three intensities (except for W2L-E which can only generate videos with happy or sad expressions in one intensity). We select five videos for each actor and compute the FED and FID between the frames of the generated video and the corresponding video with the target expression type in the MEAD dataset. The ID is computed with the input. As shown in Table \ref{tab:video}, our method achieves the lowest FED and FID, and the highest ID, indicating that (1) our edited expressions are the closest to ground truth videos with the target expression type and (2) the identities of the characters are best preserved. Moreover, our method has the second-best LSE-D and LSE-C scores. We note that there is a trade-off between the ability to express emotions and the preservation of the original mouth shape, and it is worthwhile to have a reasonable deformation of the mouth, which is essential to express some strong emotions such as happiness, anger and surprise. Although the lip synchronization of our method is slightly worse than StarGAN, it is still very close to the input, and the trade-off gives our method a better expressiveness ability for emotions with strong intensity. In addition, this trade-off is controllable by adjusting the weight of $\mathcal L_{mouth}$. Our method also has the highest CPBD among all the methods, and even higher CPBD can be easily achieved by increasing the resolution of the StyleGAN.

\subsubsection{User Study}
To further examine the effectiveness of our method, we use different methods to generate 14 videos covering all seven facial expression types and four actors, and perform a user study in which 20 participants (12 males and 8 females, average age 26 years and SD = 3.94 years) are invited and asked to rate the results in terms of various metrics, including the accuracy of the generated expression, identity preservation, audio-visual synchronization, inter-frame continuity and overall quality. Table \ref{tab:user_study} summarizes the average ratings, where our method outperforms other methods in all aspects.

\begin{figure}[!ht]
\centering
\includegraphics[width=0.45\textwidth]{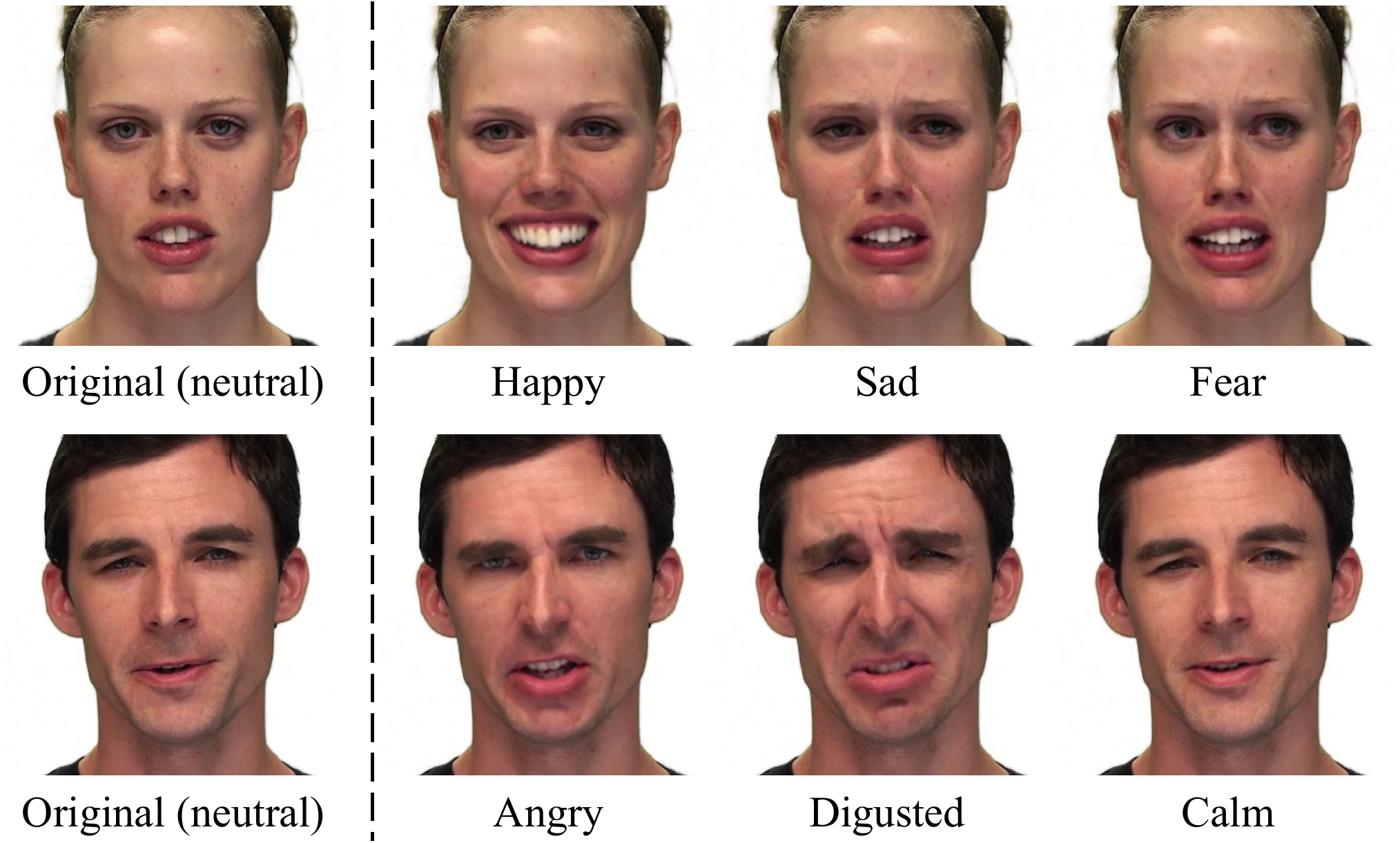}
\caption{Editing results (right) on the RAVDESS \cite{livingstone18ravdess} dataset. The input (left) is an image with a neutral emotion and the intensity of the target emotion is set to 1 (the highest intensity).}
\label{fig:emotions_ravdess}
\end{figure}

\begin{figure}[!ht]
\centering
\includegraphics[width=0.48\textwidth]{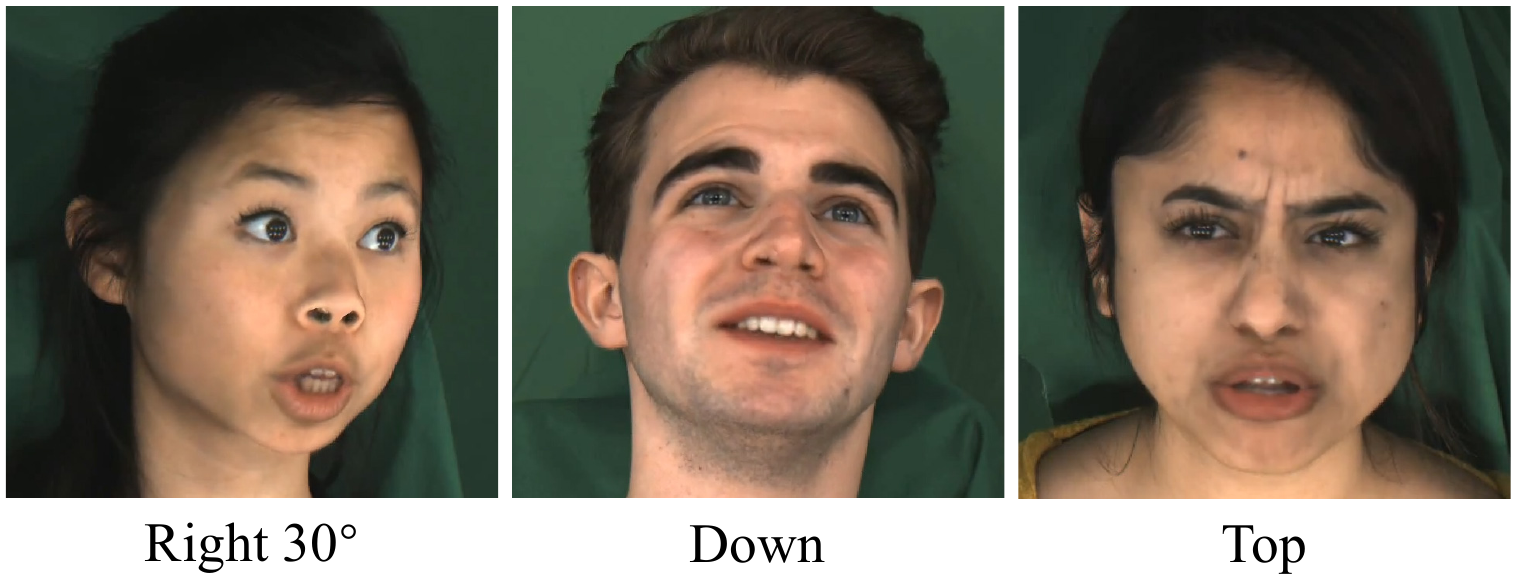}
\caption{Our method can edit faces with poses unseen in the training data.}
\label{fig:pose}
\end{figure}

\subsection{Additional Results}
\label{sec:additional_results}
We provide more animated editing results and comparisons in the supplemental demo videos. In addition to the MEAD dataset, we further conduct experiments on another dataset named RAVDESS \cite{livingstone18ravdess}. Even though the dataset is about one-tenth the size of MEAD, we still achieve relatively good editing results, as shown in Figure \ref{fig:emotions_ravdess} and the demo video. This suggests that our method has potential in adapting to new datasets.

We also note that our method uses 3DMM to decouple face shape from pose and scale, thus making our method robust to changes in pose and scale. This even enables our method to edit faces with those poses unseen in the training data, as shown in Figure \ref{fig:pose} and the demo video.

\section{Conclusion}
In this paper, we propose a simple yet effective method for high-quality facial expression editing in talking videos. Our method can output videos with specified target emotions of continuous intensities. Our key idea is regarding facial expression editing in talking videos as a special motion information editing, and we propose a two-level transformation strategy. This strategy uses a 3DMM to capture major facial movements (level-1) and an associated texture map modeled by StyleGAN to capture appearance details (level-2). Extensive experiments, including qualitative and quantitative evaluations, an ablation study and a user study show the effectiveness of this strategy and demonstrate that our method outperforms existing methods and achieves a good trade-off in terms of commonly used quality measures such as FID, FED, CPBD, ID, LSE-D and LSE-C, as well as a new proposed metric LIE. In the future work, we plan to explore the feasibility of training a universal model capable of generating high-quality results for multiple people.


%



\ifCLASSOPTIONcompsoc
  \section*{Acknowledgments}
\else
  \section*{Acknowledgment}
\fi

We would like to thank the authors of EVP \cite{ji2021audio} for providing us with relevant experimental results.

\ifCLASSOPTIONcaptionsoff
  \newpage
\fi



\bibliographystyle{IEEEtran}
\bibliography{IEEEabrv,main}
%



%




\appendix
\counterwithin{figure}{section}  
\counterwithin{table}{section}
\setcounter{figure}{0}
\setcounter{table}{0}

\section*{Training and Inference Speed}

All experiments are conducted on a server with the Intel Xeon Gold 6126 CPU and NVIDIA Titan RTX GPUs. Training two StyleGANs takes 28 hours on four GPUs and training two corresponding encoders and a shape transformation network takes 20 hours on one GPU. During inference, we generate 1080p videos at a rate of about 6 fps. However, we believe this rate can be further improved by utilizing batch processing instead of processing each frame individually.

\section*{Multi-Label Emotion Editing}

\begin{figure}[!ht]
\centering
\includegraphics[width=0.48\textwidth]{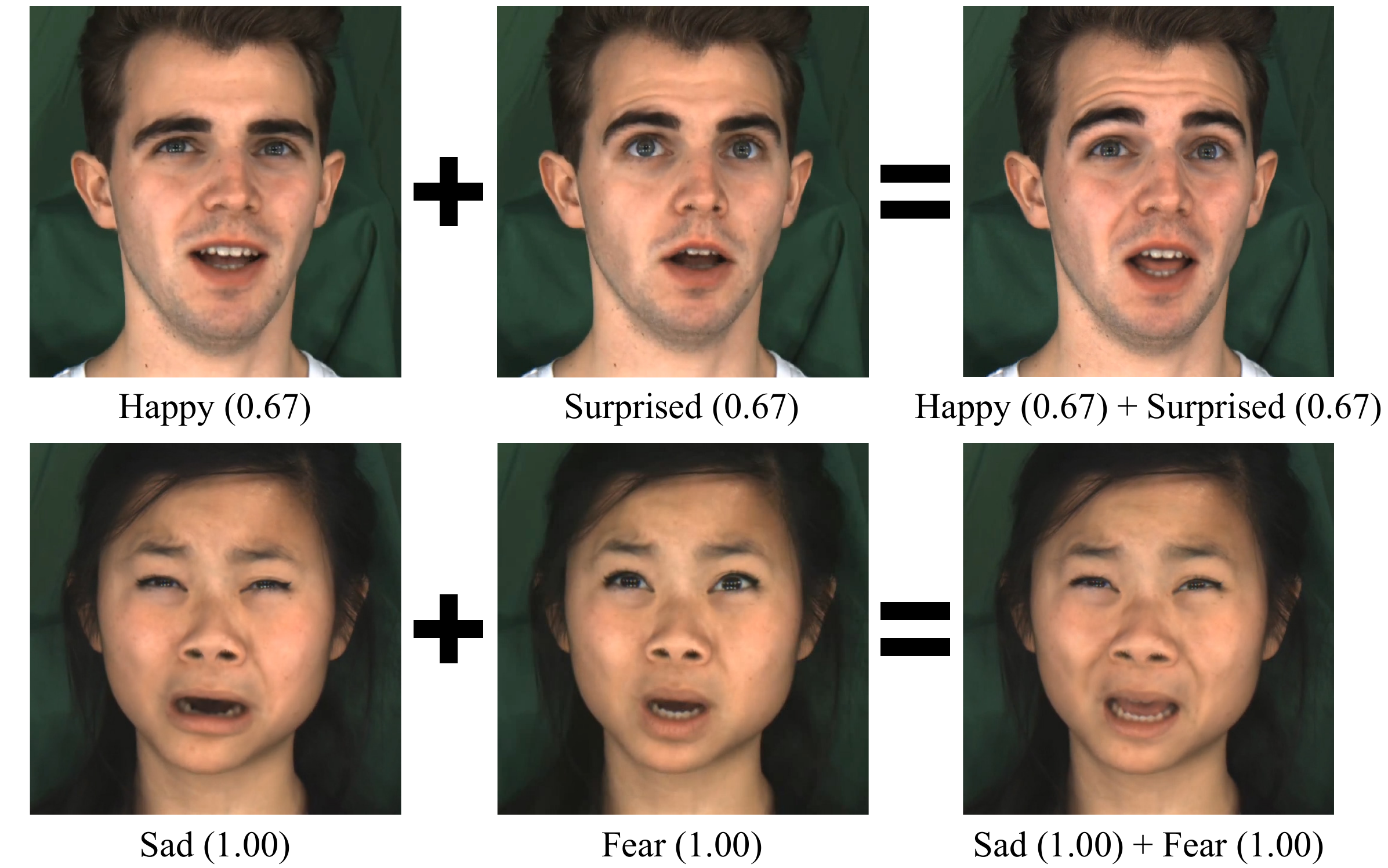}
\caption{Our method has a certain ability to generate multi-label emotions.}
\label{fig:multiple_emotions}
\end{figure}

Our method is specifically designed for generating a single emotion. However, we find it has a certain ability to generate multi-label emotions. We present some results in Fig.~\ref{fig:multiple_emotions} and the demo video.

\section*{Editing of Audio-Driven Talking Faces}
\begin{figure}[!ht]
\centering
\includegraphics[width=0.49\textwidth]{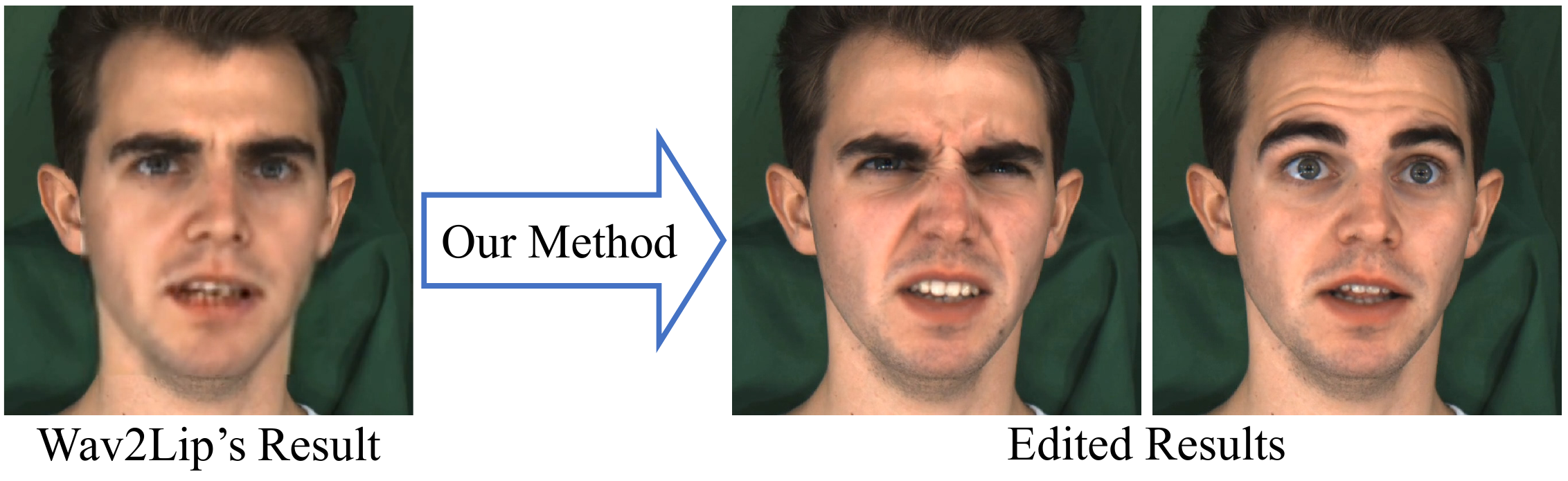}
\caption{Our method can edit the expressions and improve the visual quality of the video generated by Wav2Lip.}
\label{fig:wav2lip}
\end{figure}

Our method can also be used to edit the facial expressions and improve the visual quality of audio-driven talking face videos. We show the editing result of a video generated by Wav2Lip \citeappend{a:prajwal2020lip} in Fig.~\ref{fig:wav2lip} and the demo video.

\section*{Failure Cases of Unseen Large Poses}
\begin{figure}[!ht]
\centering
\includegraphics[width=0.45\textwidth]{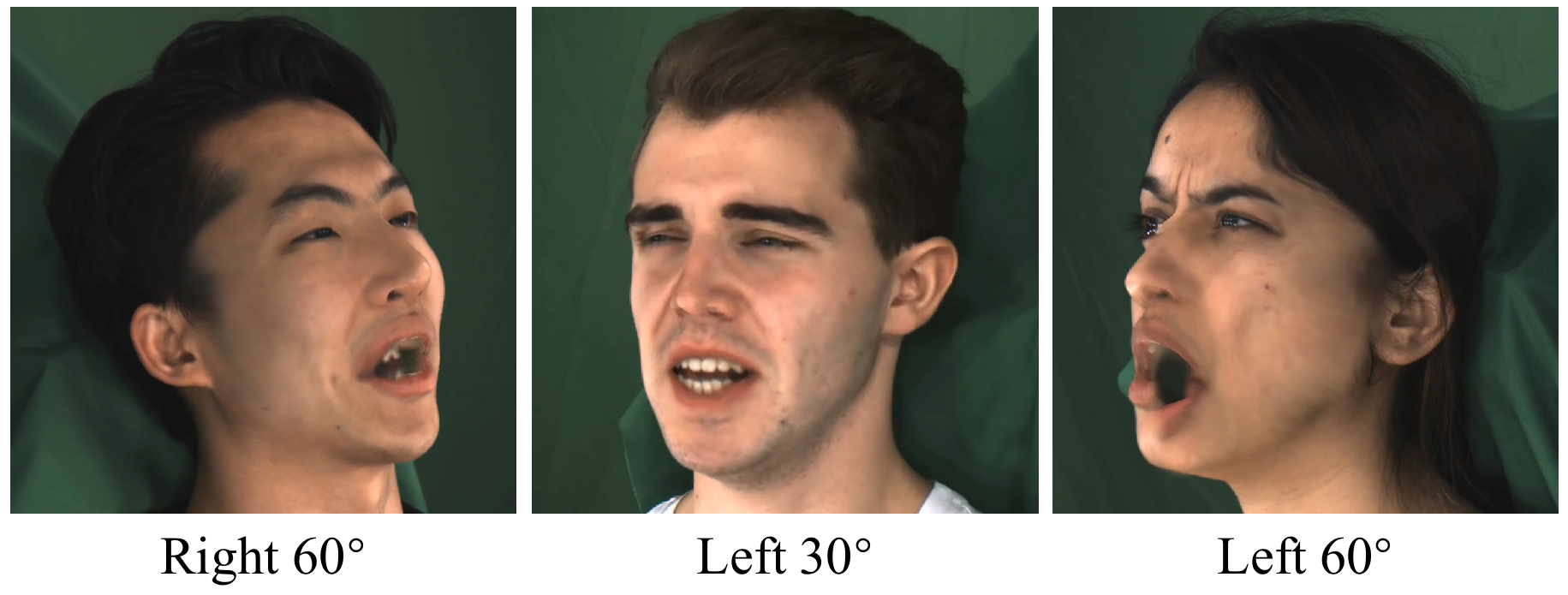}
\caption{Our method may fail when the poses differ greatly from those in the dataset.}
\label{fig:large_pose}
\end{figure}

Although our model is robust to changes in pose and scale and can handle small poses that were not seen during training, it may still fail when the poses differs greatly from those in the dataset. In such situations, the trained texture may fail to adequately cover the face or reflect the specular light from the new angle. Additionally, the projective transformation used to align the generated frontal mouth area may produce unnatural results. Fig.~\ref{fig:large_pose} illustrates some instances of these failure cases. We believe that employing neural implicit representation models like NeRF could help overcome this limitation in the future.

\section*{Using other 3DMM}

\begin{figure}[!ht]
\centering
\includegraphics[width=0.49\textwidth]{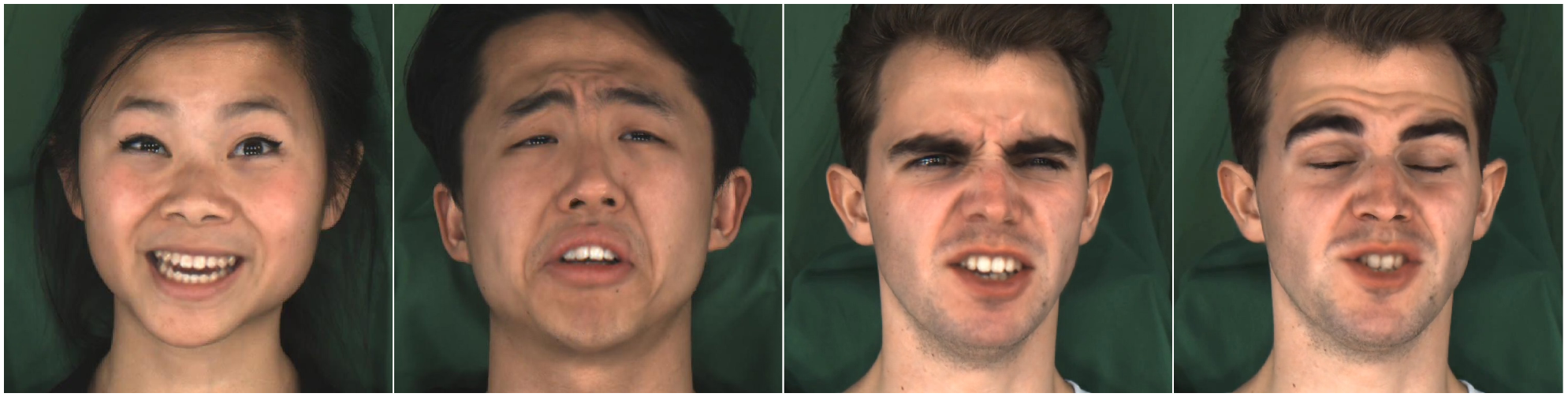}
\caption{Editing results using DECA  as the face reconstruction method.}
\label{fig:deca}
\end{figure}

We substitute our 3DMM with FLAME \citeappend{a:li2017learning} and employ DECA \citeappend{a:feng2021learning} for 3D face reconstruction. We select only the facial area of the FLAME mesh, and use the ``shape'', ``exp'' and jaw pose parameters in the shape transformation network. The other settings are kept the same as in previous experiments. We show qualitative in Fig.~\ref{fig:deca} and quantitative results in Table~\ref{tab:deca}. There are no significant differences between the two settings except that our original setting has better lip shape preservation, indicating our method can be adapted to other face models and reconstruction methods.

\begin{table}[!ht]
    \setlength\tabcolsep{4pt}
    \renewcommand{\arraystretch}{1.3}
    \caption{Quantitative evaluation of expression editing effect, lip synchronization and image quality on video editing, using DECA as the face reconstruction method.}
    \label{tab:deca}
    \centering
    \begin{tabular}{c|cccccc}
    \hline
    Metric & FED$\downarrow$ & FID$\downarrow$ & ID$\uparrow$ & LSE-D$\downarrow$ & LSE-C$\uparrow$ & CPBD$\uparrow$ \\
    \hline
    Ours (DECA) & 9.84 & 30.49  & 0.931 & 8.36 & 6.71 & 0.167 \\

    \hline
    \end{tabular}
\end{table}

\bibliographystyleappend{IEEEtran}
\bibliographyappend{IEEEabrv,appendix}


\end{document}